# Image Deblurring and Super-resolution by Adaptive Sparse Domain Selection and Adaptive Regularization

Weisheng Dong[a,b], Lei Zhang[b,1], *Member, IEEE,*
Guangming Shi[a], *Senior Member, IEEE,* and Xiaolin Wu[c], *Senior Member, IEEE*

[a]Key Laboratory of Intelligent Perception and Image Understanding (Chinese Ministry of Education), School of Electronic Engineering, Xidian University, China
[b]Dept. of Computing, The Hong Kong Polytechnic University, Hong Kong
[c]Dept. of Electrical and Computer Engineering, McMaster University, Canada

**Abstract:** As a powerful statistical image modeling technique, sparse representation has been successfully used in various image restoration applications. The success of sparse representation owes to the development of $l_1$-norm optimization techniques, and the fact that natural images are intrinsically sparse in some domain. The image restoration quality largely depends on whether the employed sparse domain can represent well the underlying image. Considering that the contents can vary significantly across different images or different patches in a single image, we propose to learn various sets of bases from a pre-collected dataset of example image patches, and then for a given patch to be processed, one set of bases are adaptively selected to characterize the local sparse domain. We further introduce two adaptive regularization terms into the sparse representation framework. First, a set of autoregressive (AR) models are learned from the dataset of example image patches. The best fitted AR models to a given patch are adaptively selected to regularize the image local structures. Second, the image non-local self-similarity is introduced as another regularization term. In addition, the sparsity regularization parameter is adaptively estimated for better image restoration performance. Extensive experiments on image deblurring and super-resolution validate that by using adaptive sparse domain selection and adaptive regularization, the proposed method achieves much better results than many state-of-the-art algorithms in terms of both PSNR and visual perception.

---

[1] Corresponding author: cslzhang@comp.polyu.edu.hk. This work is supported by the Hong Kong RGC General Research Fund (PolyU 5375/09E).

# I. Introduction

Image restoration (IR) aims to reconstruct a high quality image *x* from its degraded measurement *y*. IR is a typical ill-posed inverse problem [1] and it can be generally modeled as

$$y = DHx + \upsilon, \qquad (1)$$

where *x* is the unknown image to be estimated, *H* and *D* are degrading operators and $\upsilon$ is additive noise. When *H* and *D* are identities, the IR problem becomes denoising; when *D* is identity and *H* is a blurring operator, IR becomes deblurring; when *D* is identity and *H* is a set of random projections, IR becomes compressed sensing [2-4]; when *D* is a down-sampling operator and *H* is a blurring operator, IR becomes (single image) super-resolution. As a fundamental problem in image processing, IR has been extensively studied in the past three decades [5-20]. In this paper, we focus on deblurring and single image super-resolution.

Due to the ill-posed nature of IR, the solution to Eq. (1) with an $l_2$-norm fidelity constraint, i.e., $\hat{x} = \arg\min_{x} \|y - DHx\|_2^2$, is generally not unique. To find a better solution, prior knowledge of natural images can be used to regularize the IR problem. One of the most commonly used regularization models is the total variation (TV) model [6-7]: $\hat{x} = \arg\min_{x} \{\|y - DHx\|_2^2 + \lambda \cdot |\nabla x|_1\}$, where $|\nabla x|_1$ is the $l_1$-norm of the first order derivative of *x* and $\lambda$ is a constant. Since the TV model favors the piecewise constant image structures, it tends to smooth out the fine details of an image. To better preserve the image edges, many algorithms have been later developed to improve the TV models [17-19, 42, 45, 47].

The success of TV regularization validates the importance of good image prior models in solving the IR problems. In wavelet based image denoising [21], researchers have found that the sparsity of wavelet coefficients can serve as good prior. This reveals the fact that many types of signals, e.g., natural images, can be sparsely represented (or coded) using a dictionary of atoms, such as DCT or wavelet bases. That is, denote by $\Phi$ the dictionary, we have $x \approx \Phi\alpha$ and most of the coefficients in $\alpha$ are close to zero. With the sparsity prior, the representation of *x* over $\Phi$ can be estimated from its observation *y* by solving the following $l_0$-minimization problem: $\hat{\alpha} = \arg\min_{\alpha} \{\|y - DH\Phi\alpha\|_2^2 + \lambda \cdot \|\alpha\|_0\}$, where the $l_0$-norm counts the number of nonzero coefficients in vector $\alpha$. Once $\hat{\alpha}$ is obtained, *x* can then be estimated as $\hat{x} = \Phi\hat{\alpha}$. The $l_0$-minimization is an NP-hard combinatorial search problem, and is usually solved by greedy algorithms [48, 60]. The $l_1$-minimization, as the closest convex function to $l_0$-minimization, is then widely used as an alternative approach to solving the sparse coding problem: $\hat{\boldsymbol{\alpha}} = \arg\min_{\boldsymbol{\alpha}} \left\{ \|\boldsymbol{y} - \boldsymbol{DH\Phi\alpha}\|_2^2 + \lambda \cdot \|\boldsymbol{\alpha}\|_1 \right\}$ [60]. In addition, recent studies showed that iteratively reweighting the $l_1$-norm sparsity regularization term can lead to better IR results [59]. Sparse representation has been successfully used in various image processing applications [2-4, 13, 21-25, 32].

A critical issue in sparse representation modeling is the determination of dictionary $\boldsymbol{\Phi}$. Analytically designed dictionaries, such as DCT, wavelet, curvelet and contourlets, share the advantages of fast implementation; however, they lack the adaptivity to image local structures. Recently, there has been much effort in learning dictionaries from example image patches [13-15, 26-31, 55], leading to state-of-the-art results in image denoising and reconstruction. Many dictionary learning (DL) methods aim at learning a universal and over-complete dictionary to represent various image structures. However, sparse decomposition over a highly redundant dictionary is potentially unstable and tends to generate visual artifacts [53-54]. In this paper we propose an adaptive sparse domain selection (ASDS) scheme for sparse representation. By learning a set of compact sub-dictionaries from high quality example image patches. The example image patches are clustered into many clusters. Since each cluster consists of many patches with similar patterns, a compact sub-dictionary can be learned for each cluster. Particularly, for simplicity we use the principal component analysis (PCA) technique to learn the sub-dictionaries. For an image patch to be coded, the best sub-dictionary that is most relevant to the given patch is selected. Since the given patch can be better represented by the adaptively selected sub-dictionary, the whole image can be more accurately reconstructed than using a universal dictionary, which will be validated by our experiments.

Apart from the sparsity regularization, other regularization terms can also be introduced to further increase the IR performance. In this paper, we propose to use the piecewise autoregressive (AR) models, which are pre-learned from the training dataset, to characterize the local image structures. For each given local patch, one or several AR models can be adaptively selected to regularize the solution space. On the other hand, considering the fact that there are often many repetitive image structures in an image, we introduce a non-local (NL) self-similarity constraint served as another regularization term, which is very helpful in preserving edge sharpness and suppressing noise.

After introducing ASDS and adaptive regularizations (AReg) into the sparse representation based IR framework, we present an efficient iterative shrinkage (IS) algorithm to solve the $l_1$-minimization problem. In addition, we adaptively estimate the image local sparsity to adjust the sparsity regularization parameters. Extensive experiments on image deblurring and super-resolution show that the proposed ASDS-AReg approach can effectively reconstruct the image details, outperforming many state-of-the-art IR methods in terms of both PSNR and visual perception.

The rest of the paper is organized as follows. Section II introduces the related works. Section III presents the ASDS-based sparse representation. Section IV describes the AReg modeling. Section V summarizes the proposed algorithm. Section VI presents experimental results and Section VII concludes the paper.

## II. Related Works

It has been found that natural images can be generally coded by structural primitives, e.g., edges and line segments [61], and these primitives are qualitatively similar in form to simple cell receptive fields [62]. In [63], Olshausen *et al.* proposed to represent a natural image using a small number of basis functions chosen out of an over-complete code set. In recent years, such a sparse coding or sparse representation strategy has been widely studied to solve inverse problems, partially due to the progress of $l_0$-norm and $l_1$-norm minimization techniques [60].

Suppose that $x \in \Re^n$ is the target signal to be coded, and $\Phi = [\phi_1, \ldots, \phi_m] \in \Re^{n \times m}$ is a given dictionary of atoms (i.e., code set). The sparse coding of $x$ over $\Phi$ is to find a sparse vector $\alpha = [\alpha_1; \ldots; \alpha_m]$ (i.e., most of the coefficients in $\alpha$ are close to zero) such that $x \approx \Phi\alpha$ [49]. If the sparsity is measured as the $l_0$-norm of $\alpha$, which counts the non-zero coefficients in $\alpha$, the sparse coding problem becomes $\min_{\alpha} \|x - \Phi\alpha\|_2^2$ s.t. $\|\alpha\|_0 \leq T$, where $T$ is a scalar controlling the sparsity [55]. Alternatively, the sparse vector $\alpha$ can also be found by

$$\hat{\alpha} = \arg\min_{\alpha} \left\{ \|x - \Phi\alpha\|_2^2 + \lambda \cdot \|\alpha\|_0 \right\}, \qquad (2)$$

where $\lambda$ is a constant. Since the $l_0$-norm is non-convex, it is often replaced by either the standard $l_1$-norm or the weighted $l_1$-norm to make the optimization problem convex [3, 57, 59, 60].

An important issue of the sparse representation modeling is the choice of dictionary $\Phi$. Much effort has been made in learning a redundant dictionary from a set of example image patches [13-15, 26-31, 55]. Given a set of training image patches $S=[s_1, \ldots, s_N] \in \Re^{n \times N}$, the goal of dictionary learning (DL) is to jointly optimize the dictionary $\Phi$ and the representation coefficient matrix $\Lambda=[\alpha_1,\ldots,\alpha_N]$ such that $s_i \approx \Phi\alpha_i$ and $\|\alpha_i\|_p \leq T$, where $p = 0$ or $1$. This can be formulated by the following minimization problem:

$$(\hat{\Phi}, \hat{\Lambda}) = \arg\min_{\Phi, \Lambda} \|S - \Phi\Lambda\|_F^2 \quad \text{s.t.} \quad \|\alpha_i\|_p \leq T, \forall i, \qquad (3)$$

where $\|\cdot\|_F$ is the Frobenius norm. The above minimization problem is non-convex even when $p=1$. To make it tractable, approximation approaches, including MOD [56] and K-SVD [26], have been proposed to alternatively optimizing $\Phi$ and $\Lambda$, leading to many state-of-the-art results in image processing [14-15, 31].

Various extensions and variants of the K-SVD algorithm [27, 29-31] have been proposed to learn a universal and over-complete dictionary. However, the image contents can vary significantly across images. One may argue that a well learned over-complete dictionary $\Phi$ can sparsely code all the possible image structures; nonetheless, for each given image patch, such a "universal" dictionary $\Phi$ is neither optimal nor efficient because many atoms in $\Phi$ are irrelevant to the given local patch. These irrelevant atoms will not only reduce the computational efficiency in sparse coding but also reduce the representation accuracy.

Regularization has been used in IR for a long time to incorporate the image prior information. The widely used TV regularizations lack flexibilities in characterizing the local image structures and often generate over-smoothed results. As a classic method, the autoregressive (AR) modeling has been successfully used in image compression [33] and interpolation [34-35]. Recently the AR model was used for adaptive regularization in compressive image recovery [40]: $\min_x \sum_i \|x_i - \chi_i a_i\|_2^2$ s.t. $y = Ax$, where $\chi_i$ is the vector containing the neighboring pixels of pixel $x_i$ within the support of the AR model, and $a_i$ is the AR parameter vector. In [40], the AR models are locally computed from an initially recovered image, and they perform much better than the TV regularization in reconstructing the edge structures. However, the AR models estimated from the initially recovered image may not be robust and tend to produce the "ghost" visual artifacts. In this paper, we will propose a learning-based adaptive regularization, where the AR models are learned from high-quality training images, to increase the AR modeling accuracy.

In recent years the non-local (NL) methods have led to promising results in various IR tasks, especially in image denoising [36, 15, 39]. The mathematical framework of NL means filtering was well established by Buades *et al.* [36]. The idea of NL methods is very simple: the patches that have similar patterns can be spatially far from each other and thus we can collect them in the whole image. This NL self-similarity prior was later employed in image deblurring [8, 20] and super-resolution [41]. In [15], the NL self-similarity prior was combined with the sparse representation modeling, where the similar image patches are simultaneously coded to improve the robustness of inverse reconstruction. In this work, we will also introduce an NL self-similarity regularization term into our proposed IR framework.

## III. Sparse Representation with Adaptive Sparse Domain Selection

In this section we propose an adaptive sparse domain selection (ASDS) scheme, which learns a series of compact sub-dictionaries and assigns adaptively each local patch a sub-dictionary as the sparse domain. With ASDS, a weighted $l_1$-norm sparse representation model will be proposed for IR tasks. Suppose that $\{\boldsymbol{\Phi}_k\}$, $k=1,2,\ldots,K$, is a set of $K$ orthonormal sub-dictionaries. Let $\boldsymbol{x}$ be an image vector, and $\boldsymbol{x}_i = \boldsymbol{R}_i \boldsymbol{x}$, $i=1,2,\ldots,N$, be the $i^{th}$ patch (size: $\sqrt{n} \times \sqrt{n}$) vector of $\boldsymbol{x}$, where $\boldsymbol{R}_i$ is a matrix extracting patch $\boldsymbol{x}_i$ from $\boldsymbol{x}$. For patch $\boldsymbol{x}_i$, suppose that a sub-dictionary $\boldsymbol{\Phi}_{k_i}$ is selected for it. Then, $\boldsymbol{x}_i$ can be approximated as $\hat{\boldsymbol{x}}_i = \boldsymbol{\Phi}_{k_i} \boldsymbol{\alpha}_i$, $\|\boldsymbol{\alpha}_i\|_1 \leq T$, via sparse coding. The whole image $\boldsymbol{x}$ can be reconstructed by averaging all the reconstructed patches $\hat{\boldsymbol{x}}_i$, which can be mathematically written as [22]

$$\hat{\boldsymbol{x}} = \left( \sum_{i=1}^{N} \boldsymbol{R}_i^T \boldsymbol{R}_i \right)^{-1} \sum_{i=1}^{N} \left( \boldsymbol{R}_i^T \boldsymbol{\Phi}_{k_i} \boldsymbol{\alpha}_i \right). \quad (4)$$

In Eq. (4), the matrix to be inverted is a diagonal matrix, and hence the calculation of Eq. (4) can be done in a pixel-by-pixel manner [22]. Obviously, the image patches can be overlapped to better suppress noise [22, 15] and block artifacts. For the convenience of expression, we define the following operator "∘":

$$\hat{\boldsymbol{x}} = \boldsymbol{\Phi} \circ \boldsymbol{\alpha} \triangleq \left( \sum_{i=1}^{N} \boldsymbol{R}_i^T \boldsymbol{R}_i \right)^{-1} \sum_{i=1}^{N} \left( \boldsymbol{R}_i^T \boldsymbol{\Phi}_{k_i} \boldsymbol{\alpha}_i \right), \quad (5)$$

where $\boldsymbol{\Phi}$ is the concatenation of all sub-dictionaries $\{\boldsymbol{\Phi}_k\}$ and $\boldsymbol{\alpha}$ is the concatenation of all $\boldsymbol{\alpha}_i$.

Let $\boldsymbol{y} = \boldsymbol{DHx} + \boldsymbol{v}$ be the observed degraded image, our goal is to recover the original image $\boldsymbol{x}$ from $\boldsymbol{y}$. With ASDS and the definition in Eq. (5), the IR problem can be formulated as follows:

$$\hat{\boldsymbol{\alpha}} = \arg \min_{\boldsymbol{\alpha}} \left\{ \|\boldsymbol{y} - \boldsymbol{DH\Phi} \circ \boldsymbol{\alpha}\|_2^2 + \lambda \|\boldsymbol{\alpha}\|_1 \right\}. \quad (6)$$

Clearly, one key procedure in the proposed ASDS scheme is the determination of $\boldsymbol{\Phi}_{k_i}$ for each local patch.

To facilitate the sparsity-based IR, we propose to learn offline the sub-dictionaries $\{\Phi_k\}$, and select online from $\{\Phi_k\}$ the best fitted sub-dictionary to each patch $x_i$.

*A. Learning the sub-dictionaries*

In order to learn a series of sub-dictionaries to code the various local image structures, we need to first construct a dataset of local image patches for training. To this end, we collected a set of high-quality natural images, and cropped from them a rich amount of image patches with size $\sqrt{n} \times \sqrt{n}$. A cropped image patch, denoted by $s_i$, will be involved in DL if its intensity variance $Var(s_i)$ is greater than a threshold $\Delta$, i.e., $Var(s_i) > \Delta$. This patch selection criterion is to exclude the smooth patches from training and guarantee that only the meaningful patches with a certain amount of edge structures are involved in DL.

Suppose that $M$ image patches $S=[s_1, s_2, \ldots, s_M]$ are selected. We aim to learn $K$ compact sub-dictionaries $\{\Phi_k\}$ from $S$ so that for each given local image patch, the most suitable sub-dictionary can be selected. To this end, we cluster the dataset $S$ into $K$ clusters, and learn a sub-dictionary from each of the $K$ clusters. Apparently, the $K$ clusters are expected to represent the $K$ distinctive patterns in $S$. To generate perceptually meaningful clusters, we perform the clustering in a feature space. In the hundreds of thousands patches cropped from the training images, many patches are approximately the rotated version of the others. Hence we do not need to explicitly make the training dataset invariant to rotation because it is naturally (nearly) rotation invariant. Considering the fact that human visual system is sensitive to image edges, which convey most of the semantic information of an image, we use the high-pass filtering output of each patch as the feature for clustering. It allows us to focus on the edges and structures of image patches, and helps to increase the accuracy of clustering. The high-pass filtering is often used in low-level statistical learning tasks to enhance the meaningful features [50].

Denote by $S_h = [s_1^h, s_2^h, \ldots, s_M^h]$ the high-pass filtered dataset of $S$. We adopt the $K$-means algorithm to partition $S_h$ into $K$ clusters $\{C_1, C_2, \cdots, C_K\}$ and denote by $\mu_k$ the centroid of cluster $C_k$. Once $S_h$ is partitioned, dataset $S$ can then be clustered into $K$ subsets $S_k$, $k=1,2,\ldots,K$, and $S_k$ is a matrix of dimension $n \times m_k$, where $m_k$ denotes the number of samples in $S_k$.

Now the remaining problem is how to learn a sub-dictionary $\Phi_k$ from the cluster $S_k$ such that all the elements in $S_k$ can be faithfully represented by $\Phi_k$. Meanwhile, we hope that the representation of $S_k$ over $\Phi_k$ is as sparse as possible. The design of $\Phi_k$ can be intuitively formulated by the following objective function:

$$(\hat{\Phi}_k, \hat{\Lambda}_k) = \arg\min_{\Phi_k, \Lambda_k} \left\{ \|S_k - \Phi_k \Lambda_k\|_F^2 + \lambda \|\Lambda_k\|_1 \right\}, \tag{7}$$

where $\Lambda_k$ is the representation coefficient matrix of $S_k$ over $\Phi_k$. Eq. (7) is a joint optimization problem of $\Phi_k$ and $\Lambda_k$, and it can be solved by alternatively optimizing $\Phi_k$ and $\Lambda_k$, like in the K-SVD algorithm [26].

However, we do not directly use Eq. (7) to learn the sub-dictionary $\Phi_k$ based on the following considerations. First, the $l_2$-$l_1$ joint minimization in Eq. (7) requires much computational cost. Second and more importantly, by using the objective function in Eq. (7) we often assume that the dictionary $\Phi_k$ is over-complete. Nonetheless, here $S_k$ is a sub-dataset after K-means clustering, which implies that not only the number of elements in $S_k$ is limited, but also these elements tend to have similar patterns. Therefore, it is not necessary to learn an over-complete dictionary $\Phi_k$ from $S_k$. In addition, a compact dictionary will decrease much the computational cost of the sparse coding of a given image patch. With the above considerations, we propose to learn a compact dictionary while trying to approximate Eq. (7). The principal component analysis (PCA) is a good solution to this end.

PCA is a classical signal de-correlation and dimensionality reduction technique that is widely used in pattern recognition and statistical signal processing [37]. In [38-39], PCA has been successfully used in spatially adaptive image denoising by computing the local PCA transform of each image patch. In this paper we apply PCA to each sub-dataset $S_k$ to compute the principal components, from which the dictionary $\Phi_k$ is constructed. Denote by $\Omega_k$ the co-variance matrix of dataset $S_k$. By applying PCA to $\Omega_k$, an orthogonal transformation matrix $P_k$ can be obtained. If we set $P_k$ as the dictionary and let $Z_k = P_k^T S_k$, we will then have $\|S_k - P_k Z_k\|_F^2 = \|S_k - P_k P_k^T S_k\|_F^2 = 0$. In other words, the approximation term in Eq. (7) will be exactly zero, yet the corresponding sparsity regularization term $\|Z_k\|_1$ will have a certain amount because all the representation coefficients in $Z_k$ are preserved.

To make a better balance between the $l_1$-norm regularization term and $l_2$-norm approximation term in Eq. (7), we only extract the first $r$ most important eigenvectors in $P_k$ to form a dictionary $\Phi_r$, i.e. $\Phi_r = [p_1, p_2, ..., p_r]$. Let $\Lambda_r = \Phi_r^T S_k$. Clearly, since not all the eigenvectors are used to form $\Phi_r$, the reconstruction error $\|S_k - \Phi_r \Lambda_r\|_F^2$ in Eq. (7) will increase with the decrease of $r$. However, the term $\|\Lambda_r\|_1$ will decrease. Therefore, the optimal value of $r$, denoted by $r_o$, can be determined by

$$r_o = \arg\min_r \left\{ \|\mathbf{S}_k - \boldsymbol{\Phi}_r \boldsymbol{\Lambda}_r\|_F^2 + \lambda \|\boldsymbol{\Lambda}_r\|_1 \right\}. \tag{8}$$

Finally, the sub-dictionary learned from sub-dataset $\mathbf{S}_k$ is $\boldsymbol{\Phi}_k = [\mathbf{p}_1, \mathbf{p}_2, ..., \mathbf{p}_{r_o}]$.

Applying the above procedures to all the $K$ sub-datasets $\mathbf{S}_k$, we could get $K$ sub-dictionaries $\boldsymbol{\Phi}_k$, which will be used in the adaptive sparse domain selection process of each given image patch. In Fig. 1, we show some example sub-dictionaries learned from a training dataset. The left column shows the centroids of some sub-datasets after $K$-means clustering, and the right eight columns show the first eight atoms in the sub-dictionaries learned from the corresponding sub-datasets.

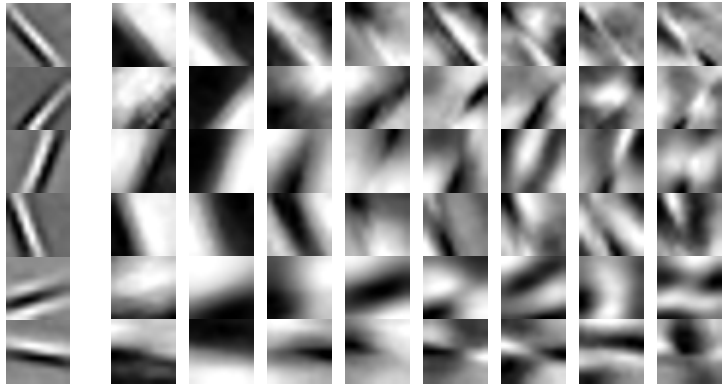

**Fig. 1.** Examples of learned sub-dictionaries. The left column shows the centroids of some sub-datasets after $K$-means clustering, and the right eight columns show the first eight atoms of the learned sub-dictionaries from the corresponding sub-datasets.

## B. Adaptive selection of the sub-dictionary

In the previous subsection, we have learned a dictionary $\boldsymbol{\Phi}_k$ for each subset $\mathbf{S}_k$. Meanwhile, we have computed the centroid $\boldsymbol{\mu}_k$ of each cluster $C_k$ associated with $\mathbf{S}_k$. Therefore, we have $K$ pairs $\{\boldsymbol{\Phi}_k, \boldsymbol{\mu}_k\}$, with which the ASDS of each given image patch can be accomplished.

In the proposed sparsity-based IR scheme, we assign adaptively a sub-dictionary to each local patch of $\mathbf{x}$, spanning the adaptive sparse domain. Since $\mathbf{x}$ is unknown beforehand, we need to have an initial estimation of it. The initial estimation of $\mathbf{x}$ can be accomplished by taking wavelet bases as the dictionary and then solving Eq. (6) with the iterated shrinkage algorithm in [10]. Denote by $\hat{\mathbf{x}}$ the estimate of $\mathbf{x}$, and denote by $\hat{\mathbf{x}}_i$ a local patch of $\hat{\mathbf{x}}$. Recall that we have the centroid $\boldsymbol{\mu}_k$ of each cluster available, and hence we could select the best fitted sub-dictionary to $\hat{\mathbf{x}}_i$ by comparing the high-pass filtered patch of $\hat{\mathbf{x}}_i$, denoted by $\hat{\mathbf{x}}_i^h$, to the centroid $\mu_k$. For example, we can select the dictionary for $\hat{x}_i$ based on the minimum distance between $\hat{x}_i^h$ and $\mu_k$, i.e.

$$k_i = \arg\min_k \left\| \hat{x}_i^h - \mu_k \right\|_2. \tag{9}$$

However, directly calculating the distance between $\hat{x}_i^h$ and $\mu_k$ may not be robust enough because the initial estimate $\hat{x}$ can be noisy. Here we propose to determine the sub-dictionary in the subspace of $\mu_k$. Let $U = [\mu_1, \mu_2, ..., \mu_K]$ be the matrix containing all the centroids. By applying SVD to the co-variance matrix of $U$, we can obtain the PCA transformation matrix of $U$. Let $\Phi_c$ be the projection matrix composed by the first several most significant eigenvectors. We compute the distance between $\hat{x}_i^h$ and $\mu_k$ in the subspace spanned by $\Phi_c$:

$$k_i = \arg\min_k \left\| \Phi_c \hat{x}_i^h - \Phi_c \mu_k \right\|_2. \tag{10}$$

Compared with Eq. (9), Eq. (10) can increase the robustness of adaptive dictionary selection.

By using Eq. (10), the $k_i^{th}$ sub-dictionary $\Phi_{k_i}$ will be selected and assigned to patch $\hat{x}_i$. Then we can update the estimation of $x$ by minimizing Eq. (6) and letting $\hat{x} = \Phi \circ \hat{\alpha}$. With the updated estimate $\hat{x}$, the ASDS of $x$ can be consequently updated. Such a process is iteratively implemented until the estimation $\hat{x}$ converges.

## C. Adaptively reweighted sparsity regularization

In Eq. (6), the parameter $\lambda$ is a constant to weight the $l_1$-norm sparsity regularization term $\|\alpha\|_1$. In [59] Candes *et al.* showed that the reweighted $l_1$-norm sparsity can more closely resemble the $l_0$-norm sparsity than using a constant weight, and consequently improve the reconstruction of sparse signals. In this sub-section, we propose a new method to estimate adaptively the image local sparsity, and then reweight the $l_1$-norm sparsity in the ASDS scheme.

The reweighted $l_1$-norm sparsity regularized minimization with ASDS can be formulated as follows:

$$\hat{\alpha} = \arg\min_\alpha \left\{ \left\| y - DH\Phi \circ \alpha \right\|_2^2 + \sum_{i=1}^{N} \sum_{j=1}^{n} \lambda_{i,j} \left| \alpha_{i,j} \right| \right\}, \tag{11}$$

where $\alpha_{i,j}$ is the coefficient associated with the $j^{th}$ atom of $\Phi_{k_i}$ and $\lambda_{i,j}$ is the weight assigned to $\alpha_{i,j}$. In [59], $\lambda_{i,j}$ is empirically computed as $\lambda_{i,j} = 1/(|\hat{\alpha}_{i,j}| + \varepsilon)$, where $\hat{\alpha}_{i,j}$ is the estimate of $\alpha_{i,j}$ and $\varepsilon$ is a small constant. Here, we propose a more robust method for computing $\lambda_{i,j}$ by formulating the sparsity estimation as a *Maximum a Posterior* (MAP) estimation problem. Under the Bayesian framework, with the observation $y$ the MAP estimation of $\alpha$ is given by

$$\hat{\alpha} = \arg\max_{\alpha} \{\log P(\alpha | y)\} = \arg\min_{\alpha} \{-\log P(y | \alpha) - \log P(\alpha)\} . \tag{12}$$

By assuming $y$ is contaminated with additive Gaussian white noises of standard deviation $\sigma_n$, we have:

$$P(y | \alpha) = \frac{1}{\sigma_n \sqrt{2\pi}} \exp(-\frac{1}{2\sigma_n^2} \|y - DH\Phi \circ \alpha\|_2^2) . \tag{13}$$

The prior distribution $P(\alpha)$ is often characterized by an i.i.d. zero-mean Laplacian probability model:

$$P(\alpha) = \prod_{i=1}^{N} \prod_{j=1}^{n} \frac{1}{\sqrt{2}\sigma_{i,j}} \exp(-\frac{\sqrt{2}}{\sigma_{i,j}} |\alpha_{i,j}|) , \tag{14}$$

where $\sigma_{i,j}$ is the standard deviation of $\alpha_{i,j}$. By plugging $P(y|\alpha)$ and $P(\alpha)$ into Eq. (12), we could readily derive the desired weight in Eq. (11) as $\lambda_{i,j} = 2\sqrt{2}\sigma_n^2 / \sigma_{i,j}$. For numerical stability, we compute the weights by

$$\lambda_{i,j} = \frac{2\sqrt{2}\sigma_n^2}{\hat{\sigma}_{i,j} + \varepsilon}, \tag{15}$$

where $\hat{\sigma}_{i,j}$ is an estimate of $\sigma_{i,j}$ and $\varepsilon$ is a small constant.

Now let's discuss how to estimate $\sigma_{i,j}$. Denote by $\hat{x}_i$ the estimate of $x_i$, and by $\hat{x}_i^l$, $l=1,2,\ldots,L$, the non-local similar patches to $\hat{x}_i$. (The determination of non-local similar patches to $\hat{x}_i$ will be described in Section IV-C.) The representation coefficients of these similar patches over the selected sub-dictionary $\Phi_{k_i}$ is $\hat{\alpha}_i^l = \Phi_{k_i}^T \hat{x}_i^l$. Then we can estimate $\sigma_{i,j}$ by calculating the standard deviation of each element $\hat{\alpha}_{i,j}$ in $\hat{\alpha}_i^l$. Compared with the reweighting method in [59], the proposed adaptive reweighting method is more robust because it exploits the image nonlocal redundancy information. Based on our experimental experience, it could lead to about 0.2dB improvement in average over the reweighting method in [59] for deblurring and super-resolution under the proposed ASDS framework. The detailed algorithm to solve the reweighted $l_1$-norm sparsity regularized minimization in Eq. (11) will be presented in Section V.

# IV. Spatially Adaptive Regularization

In Section III, we proposed to select adaptively a sub-dictionary to code the given image patch. The proposed ASDS-based IR method can be further improved by introducing two types of adaptive regularization (AReg) terms. A local area in a natural image can be viewed as a stationary process, which can be well modeled by the autoregressive (AR) models. Here, we propose to learn a set of AR models from the clustered high quality training image patches, and adaptively select one AR model to regularize the input image patch. Besides the AR models, which exploit the image local correlation, we propose to use the non-local similarity constraint as a complementary AReg term to the local AR models. With the fact that there are often many repetitive image structures in natural images, the image non-local redundancies can be very helpful in image enhancement.

## A. Training the AR models

Recall that in Section III, we have partitioned the whole training dataset into $K$ sub-datasets $S_k$. For each $S_k$ an AR model can be trained using all the sample patches inside it. Here we let the support of the AR model be a square window, and the AR model aims to predict the central pixel of the window by using the neighboring pixels. Considering that determining the best order of the AR model is not trivial, and a high order AR model may cause data over-fitting, in our experiments a 3×3 window (i.e., AR model of order 8) is used. The vector of AR model parameters, denoted by $a_k$, of the $k^{th}$ sub-dataset $S_k$, can be easily computed by solving the following least square problem:

$$a_k = \arg\min_{a} \sum_{s_i \in S_k} (s_i - a^T q_i)^2 , \qquad (16)$$

where $s_i$ is the central pixel of image patch $s_i$ and $q_i$ is the vector that consists of the neighboring pixels of $s_i$ within the support of the AR model. By applying the AR model training process to each sub-dataset, we can obtain a set of AR models $\{a_1, a_2, \ldots, a_K\}$ that will be used for adaptive regularization.

## B. Adaptive selection of the AR model for regularization

The adaptive selection of the AR model for each patch $x_i$ is the same as the selection of sub-dictionary for $x_i$ described in Section III-B. With an estimation $\hat{x}_i$ of $x_i$, we compute its high-pass Gaussian filtering output $\hat{x}_i^h$. Let $k_i = \arg\min_k \left\| \boldsymbol{\Phi}_c \hat{x}_i^h - \boldsymbol{\Phi}_c \boldsymbol{\mu}_k \right\|_2$, and then the $k_i^{\text{th}}$ AR model $\boldsymbol{a}_{k_i}$ will be assigned to patch $\boldsymbol{x}_i$. Denote by $x_i$ the central pixel of patch $\boldsymbol{x}_i$, and by $\boldsymbol{\chi}_i$ the vector containing the neighboring pixels of $x_i$ within patch $\boldsymbol{x}_i$. We can expect that the prediction error of $x_i$ using $\boldsymbol{a}_{k_i}$ and $\boldsymbol{\chi}_i$ should be small, i.e., $\left\| x_i - \boldsymbol{a}_{k_i}^T \boldsymbol{\chi}_i \right\|_2^2$ should be minimized. By incorporating this constraint into the ASDS based sparse representation model in Eq. (11), we have a lifted objective function as follows:

$$\hat{\boldsymbol{\alpha}} = \arg\min_{\boldsymbol{\alpha}} \left\{ \left\| \boldsymbol{y} - \boldsymbol{DH\Phi} \circ \boldsymbol{\alpha} \right\|_2^2 + \sum_{i=1}^{N}\sum_{j=1}^{n} \lambda_{i,j} \left| \alpha_{i,j} \right| + \gamma \cdot \sum_{x_i \in \boldsymbol{x}} \left\| x_i - \boldsymbol{a}_{k_i}^T \boldsymbol{\chi}_i \right\|_2^2 \right\}, \qquad (17)$$

where $\gamma$ is a constant balancing the contribution of the AR regularization term. For the convenience of expression, we write the third term $\sum_{x_i \in \boldsymbol{x}} \left\| x_i - \boldsymbol{a}_{k_i}^T \boldsymbol{\chi}_i \right\|_2^2$ as $\left\| (\boldsymbol{I} - \boldsymbol{A})\boldsymbol{x} \right\|_2^2$, where $\boldsymbol{I}$ is the identity matrix and

$$A(i,j) = \begin{cases} a_i, & \text{if } x_j \text{ is an element of } \boldsymbol{\chi}_i, \, a_i \in \boldsymbol{a}_{k_i} \\ 0, & \text{otherwise} \end{cases}.$$

Then, Eq. (17) can be rewritten as

$$\hat{\boldsymbol{\alpha}} = \arg\min_{\boldsymbol{\alpha}} \left\{ \left\| \boldsymbol{y} - \boldsymbol{DH\Phi} \circ \boldsymbol{\alpha} \right\|_2^2 + \sum_{i=1}^{N}\sum_{j=1}^{n} \lambda_{i,j} \left| \alpha_{i,j} \right| + \gamma \cdot \left\| (\boldsymbol{I} - \boldsymbol{A})\boldsymbol{x} \right\|_2^2 \right\}. \qquad (18)$$

## C. Adaptive regularization by non-local similarity

The AR model based AReg exploits the local statistics in each image patch. On the other hand, there are often many repetitive patterns throughout a natural image. Such non-local redundancy is very helpful to improve the quality of reconstructed images. As a complementary AReg term to AR models, we further introduce a non-local similarity regularization term into the sparsity-based IR framework.

For each local patch $\boldsymbol{x}_i$, we search for the similar patches to it in the whole image $\boldsymbol{x}$ (in practice, in a large enough area around $\boldsymbol{x}_i$). A patch $\boldsymbol{x}_i^l$ is selected as a similar patch to $\boldsymbol{x}_i$ if $e_i^l = \| \hat{\boldsymbol{x}}_i - \hat{\boldsymbol{x}}_i^l \|_2^2 \leq t$, where $t$ is a preset threshold, and $\hat{\boldsymbol{x}}_i$ and $\hat{\boldsymbol{x}}_i^l$ are the current estimates of $\boldsymbol{x}_i$ and $\boldsymbol{x}_i^l$, respectively. Or we can select the patch $\hat{\boldsymbol{x}}_i^l$ if it is within the first $L$ ($L=10$ in our experiments) closest patches to $\hat{\boldsymbol{x}}_i$. Let $x_i$ be the central pixel of patch $\boldsymbol{x}_i$, and $x_i^l$ be the central pixel of patch $\boldsymbol{x}_i^l$. Then we can use the weighted average of $x_i^l$, i.e., $\sum_{l=1}^{L} b_i^l x_i^l$, to predict $x_i$, and the weight $b_i^l$ assigned to $x_i^l$ is set as $b_i^l = \exp(-e_i^l / h) / c_i$, where $h$ is a controlling factor of the weight and $c_i = \sum_{l=1}^{L} \exp(-e_i^l / h)$ is the normalization factor. Considering that there is much non-local redundancy in natural images, we expect that the prediction error $\left\| x_i - \sum_{l=1}^{L} b_i^l x_i^l \right\|_2^2$ should be small. Let $\boldsymbol{b}_i$ be the column vector containing all the weights $b_i^l$ and $\boldsymbol{\beta}_i$ be the column vector containing all $x_i^l$. By incorporating the non-local similarity regularization term into the ASDS based sparse representation in Eq. (11), we have:

$$\hat{\boldsymbol{\alpha}} = \arg\min_{\boldsymbol{\alpha}} \left\{ \| \boldsymbol{y} - \boldsymbol{DH\Phi} \circ \boldsymbol{\alpha} \|_2^2 + \sum_{i=1}^{N}\sum_{j=1}^{n} \lambda_{i,j} |\alpha_{i,j}| + \eta \cdot \sum_{x_i \in \boldsymbol{x}} \| x_i - \boldsymbol{b}_i^T \boldsymbol{\beta}_i \|_2^2 \right\}, \tag{19}$$

where $\eta$ is a constant balancing the contribution of non-local regularization. Eq. (19) can be rewritten as

$$\hat{\boldsymbol{\alpha}} = \arg\min_{\boldsymbol{\alpha}} \left\{ \| \boldsymbol{y} - \boldsymbol{DH\Phi} \circ \boldsymbol{\alpha} \|_2^2 + \sum_{i=1}^{N}\sum_{j=1}^{n} \lambda_{i,j} |\alpha_{i,j}| + \eta \cdot \| (\boldsymbol{I} - \boldsymbol{B})\boldsymbol{\Phi\alpha} \|^2 \right\}, \tag{20}$$

where $\boldsymbol{I}$ is the identity matrix and

$$\boldsymbol{B}(i,l) = \begin{cases} b_i^l, & \text{if } x_i^l \text{ is an element of } \boldsymbol{\beta}_i, \, b_i^l \in \boldsymbol{b}_i \\ 0, & \text{otherwise} \end{cases}.$$

## V. Summary of the Algorithm

By incorporating both the local AR regularization and the non-local similarity regularization into the ASDS based sparse representation in Eq. (11), we have the following ASDS-AReg based sparse representation to solve the IR problem:

$$\hat{\boldsymbol{\alpha}} = \arg\min_{\boldsymbol{\alpha}} \left\{ \| \boldsymbol{y} - \boldsymbol{DH\Phi} \circ \boldsymbol{\alpha} \|_2^2 + \gamma \cdot \| (\boldsymbol{I} - \boldsymbol{A})\boldsymbol{\Phi} \circ \boldsymbol{\alpha} \|_2^2 + \eta \cdot \| (\boldsymbol{I} - \boldsymbol{B})\boldsymbol{\Phi} \circ \boldsymbol{\alpha} \|_2^2 + \sum_{i=1}^{N}\sum_{j=1}^{n} \lambda_{i,j} |\alpha_{i,j}| \right\}. \tag{21}$$

In Eq. (21), the first $l_2$-norm term is the fidelity term, guaranteeing that the solution $\hat{\boldsymbol{x}} = \boldsymbol{\Phi} \circ \hat{\boldsymbol{\alpha}}$ can well fit the observation $\boldsymbol{y}$ after degradation by operators $\boldsymbol{H}$ and $\boldsymbol{D}$; the second $l_2$-norm term is the local AR model based adaptive regularization term, requiring that the estimated image is locally stationary; the third $l_2$-norm term is the non-local similarity regularization term, which uses the non-local redundancy to enhance each local patch; and the last weighted $l_1$-norm term is the sparsity penalty term, requiring that the estimated image should be sparse in the adaptively selected domain. Eq. (21) can be re-written as

$$\hat{\boldsymbol{\alpha}} = \arg\min_{\boldsymbol{\alpha}} \left\| \begin{bmatrix} \boldsymbol{y} \\ \boldsymbol{0} \\ \boldsymbol{0} \end{bmatrix} - \begin{bmatrix} \boldsymbol{DH} \\ \gamma \cdot (\boldsymbol{I} - \boldsymbol{A}) \\ \eta \cdot (\boldsymbol{I} - \boldsymbol{B}) \end{bmatrix} \boldsymbol{\Phi} \circ \boldsymbol{\alpha} \right\|_2^2 + \sum_{i=1}^{N} \sum_{j=1}^{n} \lambda_{i,j} |\alpha_{i,j}|. \tag{22}$$

By letting

$$\tilde{\boldsymbol{y}} = \begin{bmatrix} \boldsymbol{y} \\ \boldsymbol{0} \\ \boldsymbol{0} \end{bmatrix}, \quad \boldsymbol{K} = \begin{bmatrix} \boldsymbol{DH} \\ \gamma \cdot (\boldsymbol{I} - \boldsymbol{A}) \\ \eta \cdot (\boldsymbol{I} - \boldsymbol{B}) \end{bmatrix}, \tag{23}$$

Eq. (22) can be re-written as

$$\hat{\boldsymbol{\alpha}} = \arg\min_{\boldsymbol{\alpha}} \left\{ \| \tilde{\boldsymbol{y}} - \boldsymbol{K}\boldsymbol{\Phi} \circ \boldsymbol{\alpha} \|_2 + \sum_{i=1}^{N} \sum_{j=1}^{n} \lambda_{i,j} |\alpha_{i,j}| \right\}. \tag{24}$$

This is a reweighted $l_1$-minimization problem, which can be effectively solved by the iterative shrinkage algorithm [10]. We outline the iterative shrinkage algorithm for solving (24) in **Algorithm 1**.

---

**Algorithm 1** for solving Eq. (24)

---

1. Initialization:
   (a) By taking the wavelet domain as the sparse domain, we can compute an initial estimate, denoted by $\hat{\boldsymbol{x}}$, of $\boldsymbol{x}$ by using the iterated wavelet shrinkage algorithm [10];
   (b) With the initial estimate $\hat{\boldsymbol{x}}$, we select the sub-dictionary $\boldsymbol{\Phi}_{k_i}$ and the AR model $\boldsymbol{a}_i$ using Eq. (10), and calculate the non-local weight $\boldsymbol{b}_i$ for each local patch $\hat{\boldsymbol{x}}_i$;
   (c) Initialize $\boldsymbol{A}$ and $\boldsymbol{B}$ with the selected AR models and the non-local weights;
   (d) Preset $\gamma$, $\eta$, $P$, $e$ and the maximal iteration number, denoted by *Max_Iter*;
   (e) Set $k=0$.
2. Iterate on $k$ until $\|\hat{\boldsymbol{x}}^{(k)} - \hat{\boldsymbol{x}}^{(k+1)}\|_2^2 / N \leq e$ or $k \geq Max\_Iter$ is satisfied.
   (a) $\hat{\boldsymbol{x}}^{(k+1/2)} = \hat{\boldsymbol{x}}^{(k)} + \boldsymbol{K}^T(\tilde{\boldsymbol{y}} - \boldsymbol{K}\hat{\boldsymbol{x}}^{(k)}) = \hat{\boldsymbol{x}}^{(k)} + (\boldsymbol{U}\boldsymbol{y} - \boldsymbol{U}\hat{\boldsymbol{x}}^{(k)} - \boldsymbol{V}\hat{\boldsymbol{x}}^{(k)})$, where $\boldsymbol{U} = (\boldsymbol{DH})^T \boldsymbol{DH}$ and $\boldsymbol{V} = \gamma^2 (\boldsymbol{I} - \boldsymbol{A})^T (\boldsymbol{I} - \boldsymbol{A}) + \eta^2 (\boldsymbol{I} - \boldsymbol{B})^T (\boldsymbol{I} - \boldsymbol{B})$;
   (b) Compute $\boldsymbol{\alpha}^{(k+1/2)} = [\boldsymbol{\Phi}_{k_1}^T \boldsymbol{R}_1 \hat{\boldsymbol{x}}^{(k+1/2)}, \cdots, \boldsymbol{\Phi}_{k_N}^T \boldsymbol{R}_N \hat{\boldsymbol{x}}^{(k+1/2)}]$, where $N$ is the total number of image patches;
   (c) $\alpha_{i,j}^{(k+1)} = \text{soft}(\alpha_{i,j}^{(k+1/2)}, \tau_{i,j})$, where $\text{soft}(\cdot, \tau_{i,j})$ is a soft thresholding function with threshold $\tau_{i,j}$;
   (d) Compute $\hat{\boldsymbol{x}}^{(k+1)} = \boldsymbol{\Phi} \circ \boldsymbol{\alpha}^{(k+1)}$ using Eq. (5), which can be calculated by first reconstructing each image patch with $\hat{\boldsymbol{x}}_i = \boldsymbol{\Phi}_{k_i} \boldsymbol{\alpha}_i^{(k+1)}$ and then averaging all the reconstructed image patches;
   (e) If $mod(k,P)=0$, update the adaptive sparse domain of $\boldsymbol{x}$ and the matrices $\boldsymbol{A}$ and $\boldsymbol{B}$ using the improved estimate $\hat{\boldsymbol{x}}^{(k+1)}$.

---

In **Algorithm 1**, $e$ is a pre-specified scalar controlling the convergence of the iterative process, and *Max_Iter* is the allowed maximum number of iterations. The thresholds $\tau_{i,j}$ are locally computed as $\tau_{i,j} = \lambda_{i,j}/r$ [10], where $\lambda_{i,j}$ are calculated by Eq. (15) and $r$ is chosen such that $r > \|(\boldsymbol{K}\boldsymbol{\Phi})^T \boldsymbol{K}\boldsymbol{\Phi}\|_2$. Since the dictionary $\Phi_{k_i}$ varies across the image, the optimal determination of *r* for each local patch is difficult. Here, we empirically set *r*=4.7 for all the patches. *P* is a preset integer, and we only update the sub-dictionaries $\Phi_{k_i}$, the AR models $a_i$ and the weights $b_i$ in every *P* iterations to save computational cost. With the updated $a_i$ and $b_i$, *A* and *B* can be updated, and then the matrix *V* can be updated.

## VI. Experimental Results

### *A. Training datasets*

Although image contents can vary a lot from image to image, it has been found that the micro-structures of images can be represented by a small number of structural primitives (e.g., edges, line segments and other elementary features), and these primitives are qualitatively similar in form to simple cell receptive fields [61-63]. The human visual system employs a sparse coding strategy to represent images, i.e., coding a natural image using a small number of basis functions chosen out of an over-complete code set. Therefore, using the many patches extracted from several training images which are rich in edges and textures, we are able to train the dictionaries which can represent well the natural images. To illustrate the robustness of the proposed method to the training dataset, we use two different sets of training images in the experiments, each set having 5 high quality images as shown in Fig. 2. We can see that these two sets of training images are very different in contents. We use $Var(s_i) > \Delta$ with $\Delta=16$ to exclude the smooth image patches, and a total amount of 727,615 patches of size 7×7 are randomly cropped from each set of training images. (Please refer to Section VI-E for the discussion of patch size selection.)

As a clustering-based method, an important issue is the selection of the number of classes. However, the optimal selection of this number is a non-trivial task, which is subject to the bias and variance tradeoff. If the number of classes is too small, the boundaries between classes will be smoothed out and thus the distinctiveness of the learned sub-dictionaries and AR models is decreased. On the other hand, a too large number of the classes will make the learned sub-dictionaries and AR models less representative and less reliable. Based on the above considerations and our experimental experience, we propose the following simple method to find a good number of classes: we first partition the training dataset into 200 clusters, and merge those classes that contain very few image patches (i.e., less than 300 patches) to their nearest neighboring classes. More discussions and experiments on the selection of the number of classes will be made in Section VI-E.

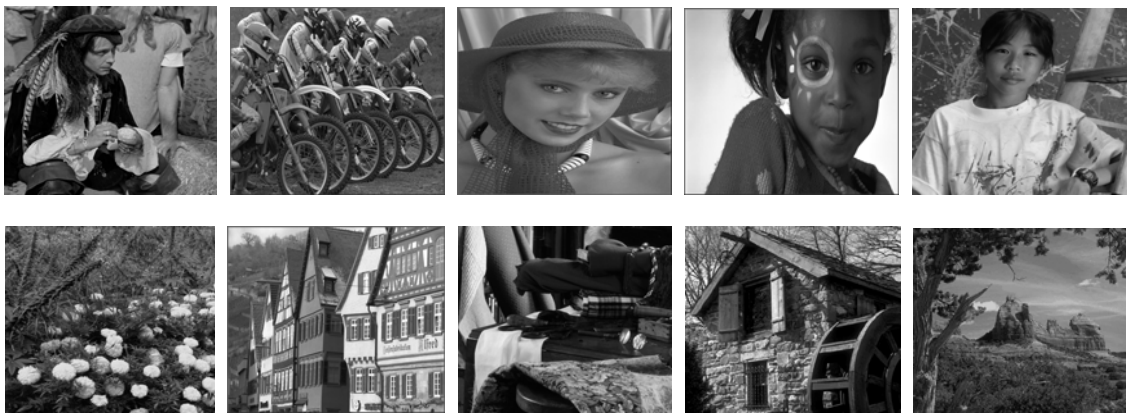

**Fig. 2.** The two sets of high quality images used for training sub-dictionaries and AR models. The images in the first row consist of the training dataset 1 and those in the second row consist of the training dataset 2.

## *B. Experimental settings*

In the experiments of deblurring, two types of blur kernels, a Gaussian kernel of standard deviation 3 and a 9×9 uniform kernel, were used to simulate blurred images. Additive Gaussian white noises with standard deviations $\sqrt{2}$ and 2 were then added to the blurred images, respectively. We compare the proposed methods with five recently proposed image deblurring methods: the iterated wavelet shrinkage method [10], the constrained TV deblurring method [42], the spatially weighted TV deblurring method [45], the $l_0$-norm sparsity based deblurring method [46], and the BM3D deblurring method [58]. In the proposed ASDS-AReg **Algorithm 1**, we empirically set $\gamma = 0.0775$, $\eta = 0.1414$, and $\tau_{i,j}=\lambda_{i,j}/4.7$, where $\lambda_{i,j}$ is adaptively computed by Eq. (15).

In the experiments of super-resolution, the degraded LR images were generated by first applying a truncated 7×7 Gaussian kernel of standard deviation 1.6 to the original image and then down-sampling by a factor of 3. We compare the proposed method with four state-of-the-art methods: the iterated wavelet shrinkage method [10], the TV-regularization based method [47], the Softcuts method [43], and the sparse representation based method [25][2]. Since the method in [25] does not handle the blurring of LR images, for fair comparisons we used the iterative back-projection method [16] to deblur the HR images produced by [25]. In the proposed ASDS-AReg based super-resolution, the parameters are set as follows. For the noiseless LR images, we empirically set $\gamma=0.0894$, $\eta=0.2$ and $\tau_{i,j}=0.18/\hat{\sigma}_{i,j}$, where $\hat{\sigma}_{i,j}$ is the estimated standard deviation of $\alpha_{i,j}$. For the noisy LR images, we empirically set $\gamma$=0.2828, $\eta$=0.5 and $\tau_{i,j}=\lambda_{i,j}$/16.6.

---

[2] We thank the authors of [42-43], [45-46], [58] and [25] for providing their source codes, executable programs, or experimental results.

In both of the deblurring and super-resolution experiments, 7×7 patches (for HR image) with 5-pixel-width overlap between adjacent patches were used in the proposed methods. For color images, all the test methods were applied to the luminance component only because human visual system is more sensitive to luminance changes, and the bi-cubic interpolator was applied to the chromatic components. Here we only report the PSNR and SSIM [44] results for the luminance component. To examine more comprehensively the proposed approach, we give three results of the proposed method: the results by using only ASDS (denoted by ASDS), by using ASDS plus AR regularization (denoted by ASDS-AR), and by using ASDS with both AR and non-local similarity regularization (denoted by ASDS-AR-NL).

## *C. Experimental results on de-blurring*

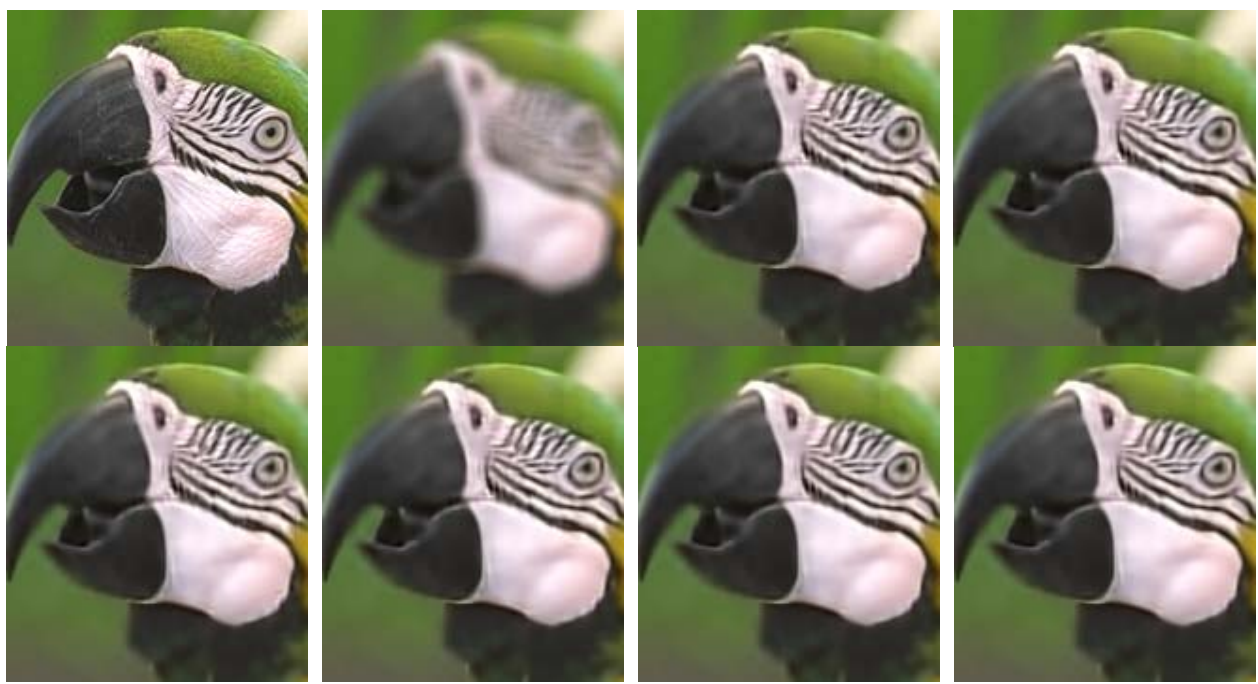

**Fig. 3.** Comparison of deblurred images (uniform blur kernel, $\sigma_n=\sqrt{2}$) on *Parrot* by the proposed methods. Top row: Original, Degraded, ASDS-TD1 (PSNR=30.71dB, SSIM=0.8926), ASDS-TD2 (PSNR=30.90dB, SSIM=0.8941). Bottom row: ASDS-AR-TD1 (PSNR=30.64dB, SSIM=0.8920), ASDS-AR-TD2 (PSNR=30.79dB, SSIM=0.8933), ASDS-AR-NL-TD1 (PSNR=30.76dB, SSIM=0.8921), ASDS-AR-NL-TD2 (PSNR=30.92dB, SSIM=0.8939).

To verify the effectiveness of ASDS and adaptive regularizations, and the robustness of them to the training datasets, we first present the deblurring results on image *Parrot* by the proposed methods in Fig. 3. More PSNR and SSIM results can be found in Table 1. From Fig. 3 and Table 1 we can see that the proposed methods generate almost the same deblurring results with TD1 and TD2. We can also see that the ASDS method is effective in deblurring. By combining the adaptive regularization terms, the deblurring results can be further improved by eliminating the ringing artifacts around edges. Due to the page limit, we will only show the results by ASDS-AR-NL-TD2 in the following development.

The deblurring results by the competing methods are then compared in Figs. 4~6. One can see that there are many noise residuals and artifacts around edges in the deblurred images by the iterated wavelet shrinkage method [10]. The TV-based methods in [42] and [45] are effective in suppressing the noises; however, they produce over-smoothed results and eliminate much image details. The $l_0$-norm sparsity based method of [46] is very effective in reconstructing smooth image areas; however, it fails to reconstruct fine image edges. The BM3D method [58] is very competitive in recovering the image structures. However, it tends to generate some "ghost" artifacts around the edges (e.g., the image *Cameraman* in Fig. 6). The proposed method leads to the best visual quality. It can not only remove the blurring effects and noise, but also reconstruct more and sharper image edges than other methods. The excellent edge preservation owes to the adaptive sparse domain selection strategy and adaptive regularizations. The PSNR and SSIM results by different methods are listed in Tables 1~4. For the experiments using uniform blur kernel, the average PSNR improvements of ASDS-AR-NL-TD2 over the second best method (i.e., BM3D [58]) are 0.50 dB (when $\sigma_n=\sqrt{2}$) and 0.4 dB (when $\sigma_n=2$), respectively. For the experiments using Gaussian blur kernel, the PSNR gaps between all the competing methods become smaller, and the average PSNR improvements of ASDS-AR-NL-TD2 over the BM3D method are 0.15 dB (when $\sigma_n=\sqrt{2}$) and 0.18 dB (when $\sigma_n=2$), respectively. We can also see that the proposed ASDS-AR-NL method achieves the highest SSIM index.

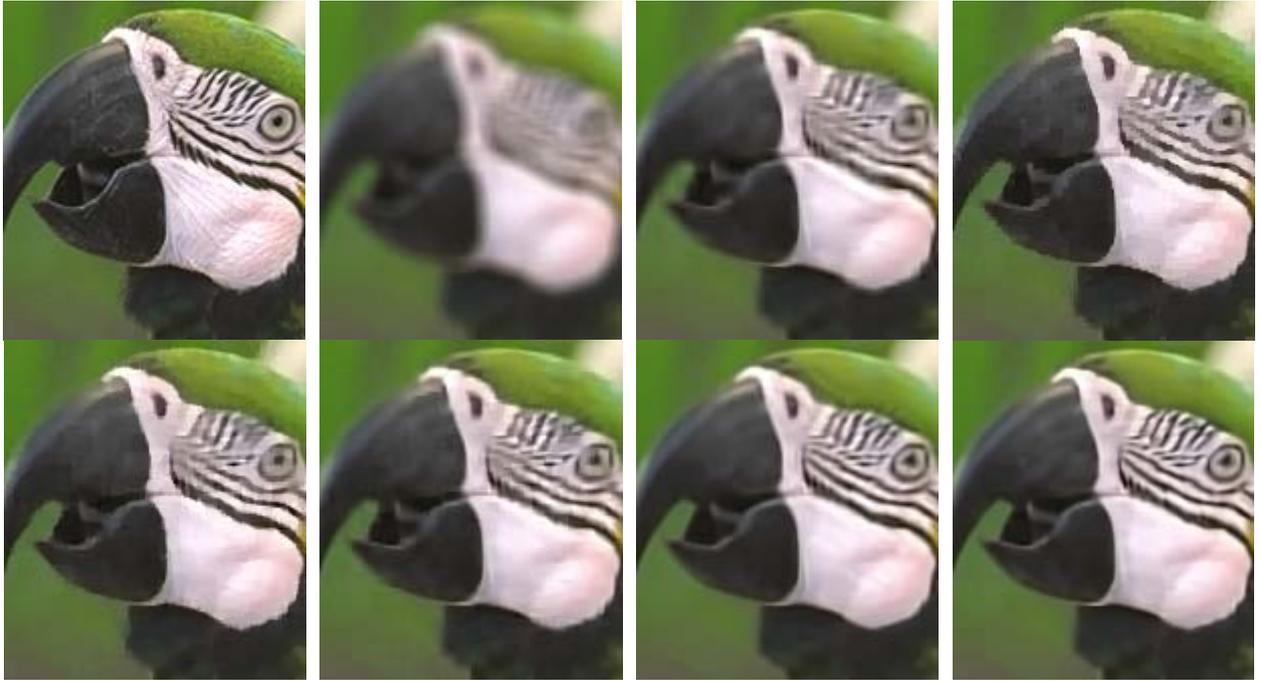

**Fig. 4.** Comparison of the deblurred images on *Parrot* by different methods (uniform blur kernel and $\sigma_n=\sqrt{2}$). Top row: Original, degraded, method [10] (PSNR=27.80dB, SSIM=0.8652) and method [42] (PSNR=28.80dB, SSIM=0.8704). Bottom row: method [45] (PSNR=28.96dB, SSIM=0.8722), method [46] (PSNR=29.04dB, SSIM=0.8824), BM3D [58] (PSNR=30.22dB, SSIM=0.8906), and proposed (PSNR=30.92dB, SSIM=0.8936).

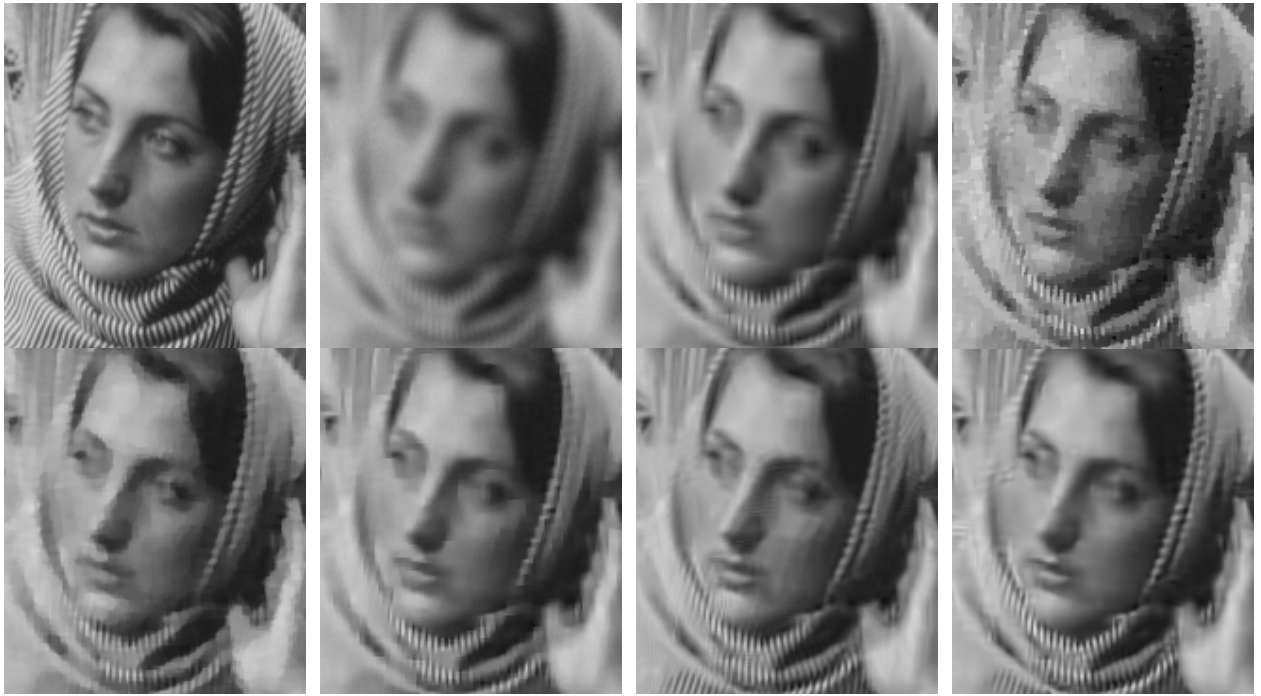

**Fig. 5.** Comparison of the deblurred images on *Barbara* by different methods (uniform blur kernel and $\sigma_n=2$). Top row: Original, degraded, method [10] (PSNR=24.86dB, SSIM=0.6963) and method [42] (PSNR=25.12dB, SSIM=0.7031). Bottom row: method [45] (PSNR=25.34dB, SSIM=0.7214), method [46] (PSNR=25.37dB, SSIM=0.7248), BM3D [58] (PSNR=27.16dB, SSIM=0.7881) and proposed (PSNR=26.96dB, SSIM=0.7927).

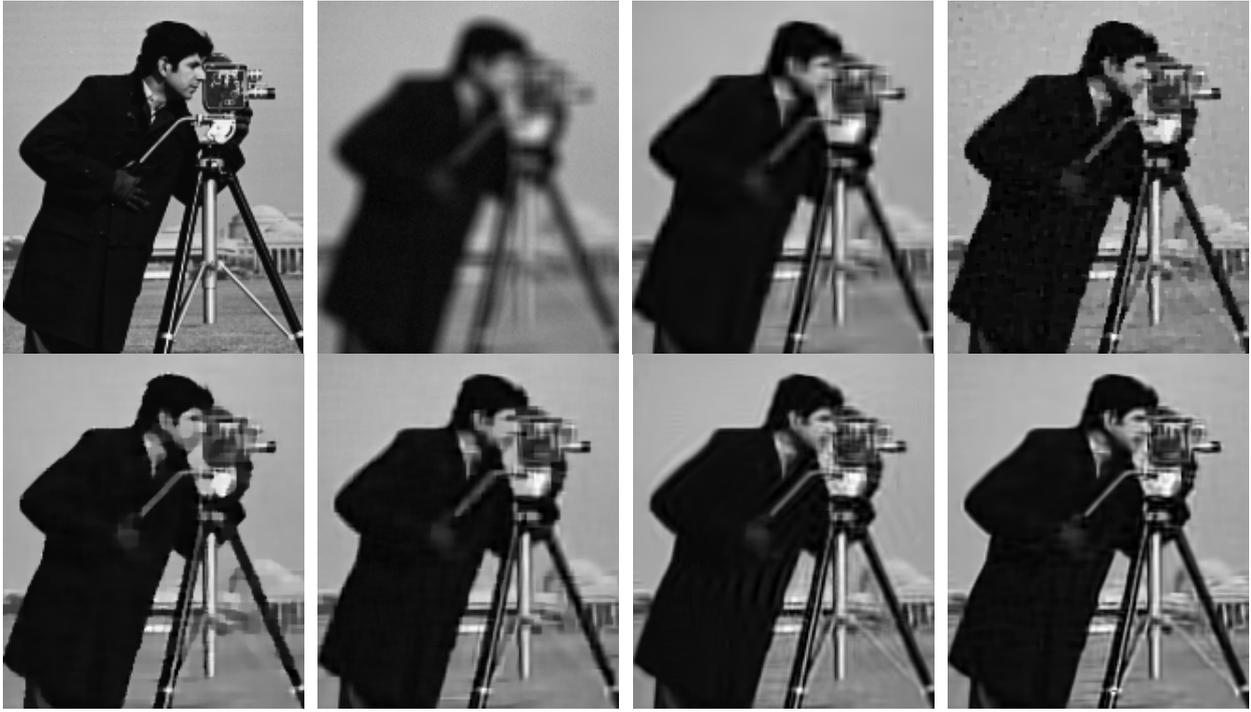

**Fig. 6.** Comparison of the deblurred images on *Cameraman* by different methods (uniform blur kernel and $\sigma_n$=2). Top row: Original, degraded, method [10] (PSNR=24.80dB, SSIM=0.7837) and method [42] (PSNR=26.04dB, SSIM=0.7772). Bottom row: method [45] (PSNR=26.53dB, SSIM=0.8273), method [46] (PSNR=25.96dB, SSIM=0.8131), BM3D [58] (PSNR=26.53 dB, SSIM=0.8136) and proposed (PSNR=27.25 dB, SSIM=0.8408).

## *D. Experimental results on single image super-resolution*

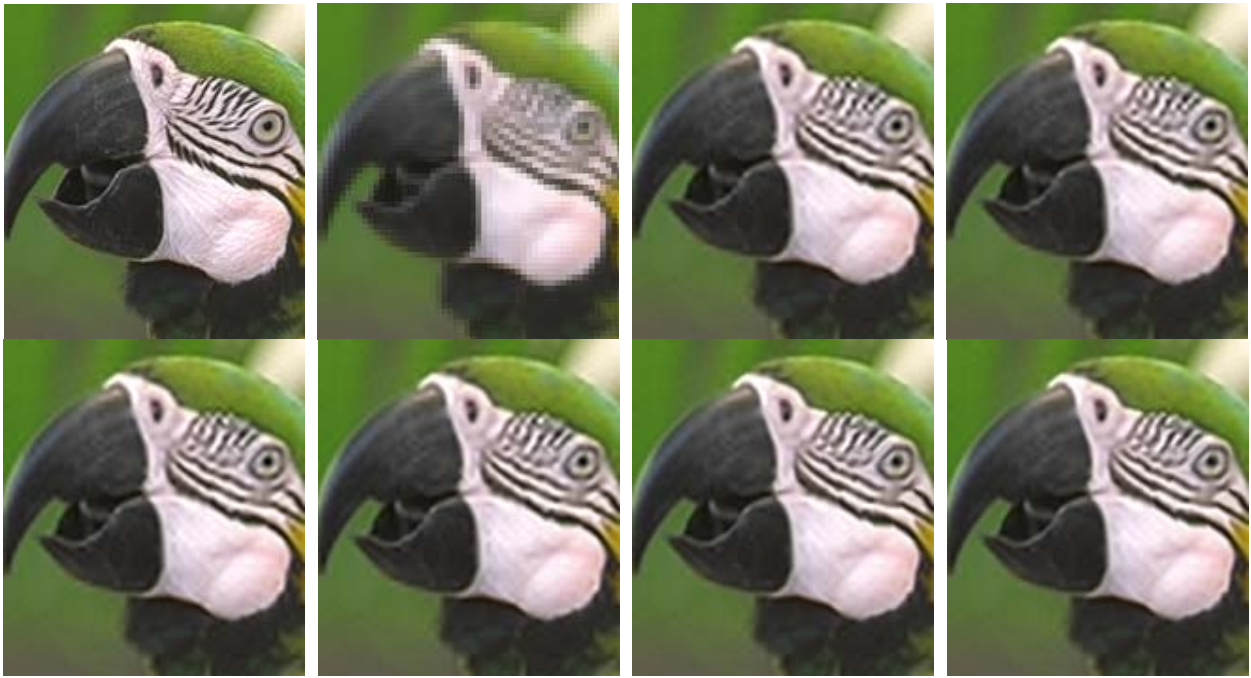

**Fig. 7**. The super-resolution results (scaling factor 3) on image *Parrot* by the proposed methods. Top row: Original, LR image, ASDS-TD1 (PSNR=29.47dB, SSIM=0.9031) and ASDS-TD2 (PSNR=29.51dB, SSIM=0.9034). Bottom row: ASDS-AR-TD1 (PSNR=29.61dB, SSIM=0.9036), ASDS-AR-TD2 (PSNR=29.63dB, SSIM=0.9038), ASDS-AR-NL-TD1 (PSNR=29.97 dB, SSIM=0.9090) and ASDS-AR-NL-TD2 (PSNR=30.00dB, SSIM=0.9093).

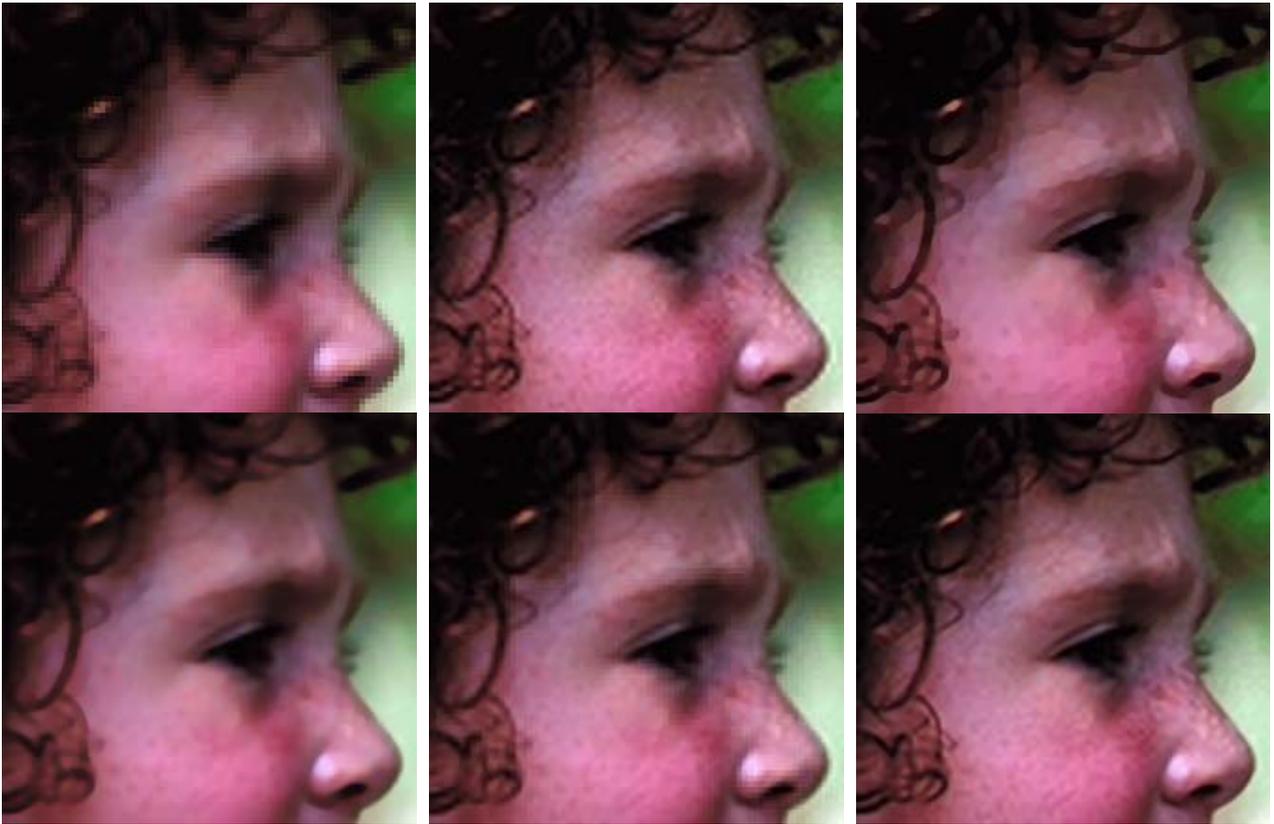

**Fig. 8.** Reconstructed HR images (scaling factor 3) of *Girl* by different methods. Top row: LR image, method [10] (PSNR=32.93dB, SSIM=0.8102) and method [47] (PSNR=31.21dB, SSIM=0.7878). Bottom row: method [43] (PSNR=31.94dB, SSIM=0.7704), method [25] (PSNR=32.51dB, SSIM=0.7912) and proposed (PSNR=33.53dB, SSIM=0.8242).

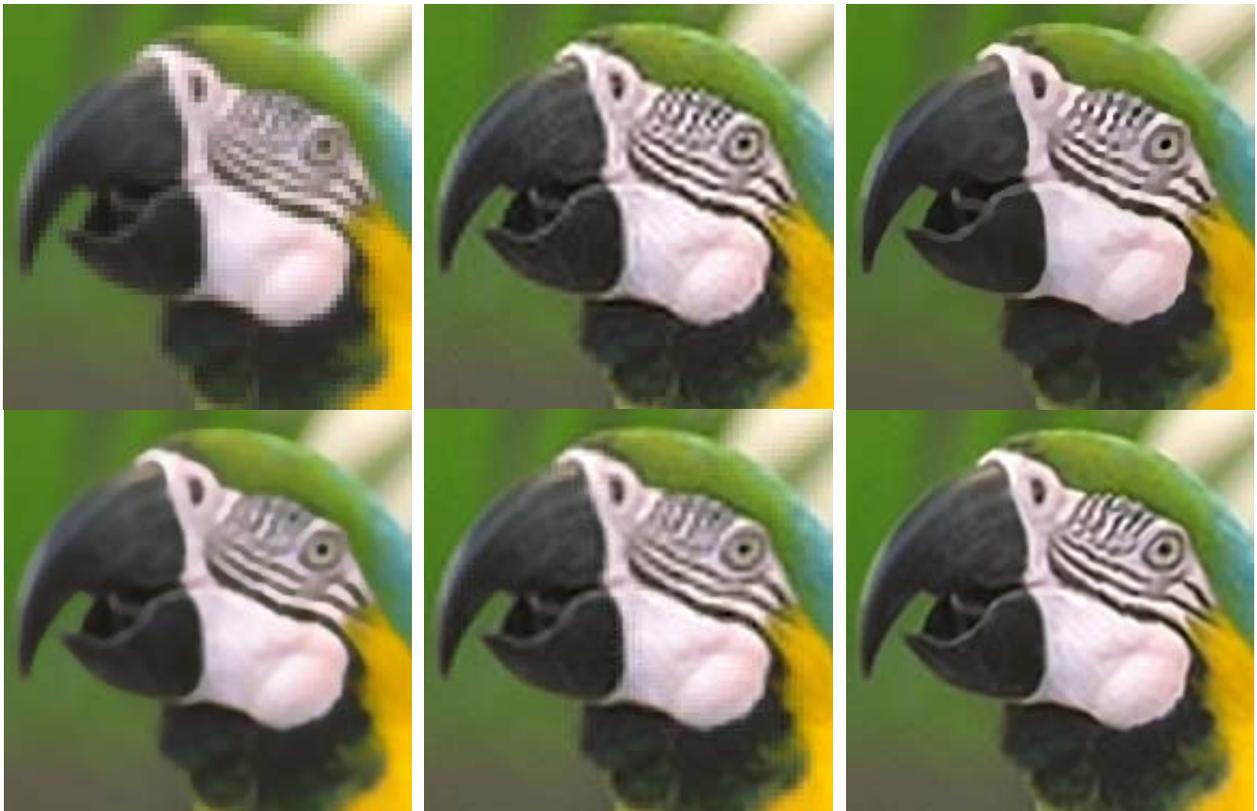

**Fig. 9.** Reconstructed HR images (scaling factor 3) of *Parrot* by different methods. Top row: LR image, method [10] (PSNR=28.78dB, SSIM=0.8845) and method [47] (PSNR=27.59dB, SSIM=0.8856). Bottom row: method [43] (PSNR=27.71dB, SSIM=0.8682), method [25] (PSNR=27.98dB, SSIM=0.8665) and proposed (PSNR=30.00dB, SSIM=0.9093).

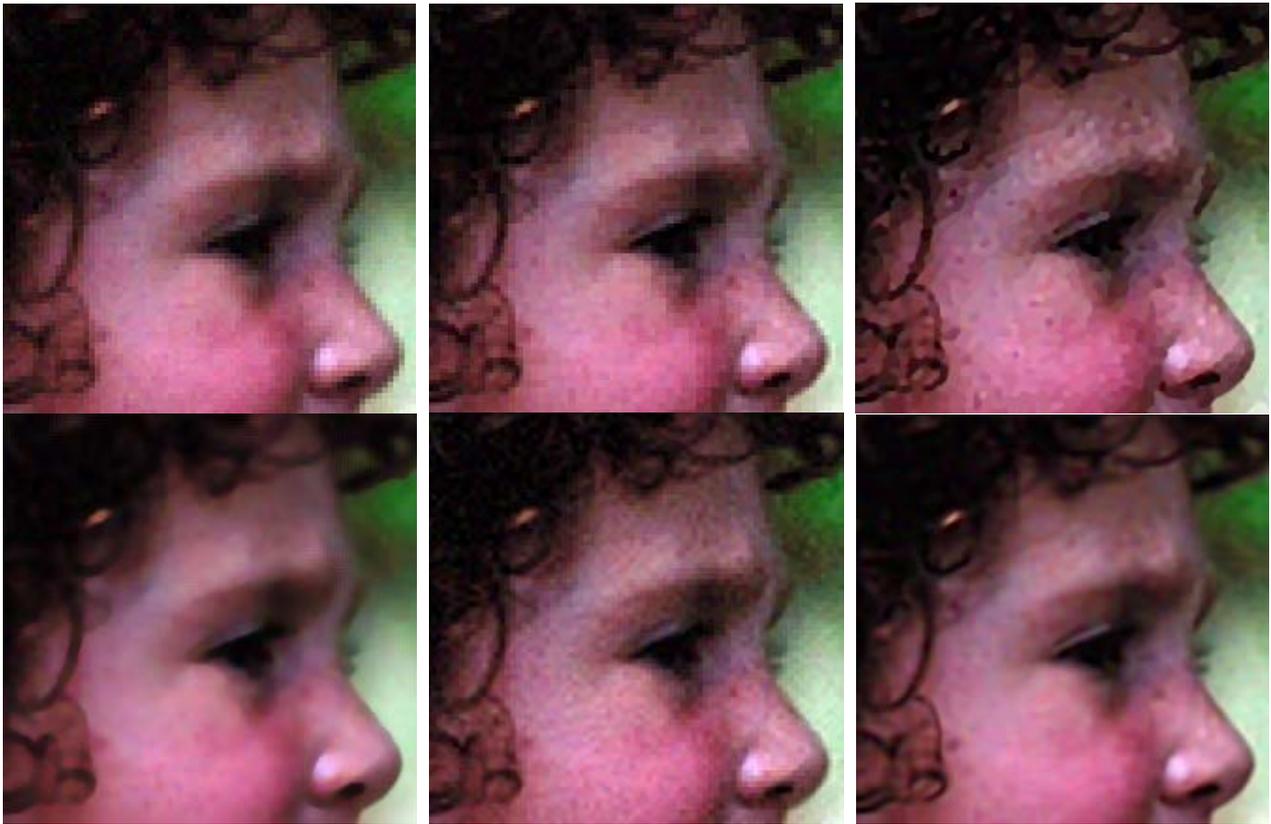

**Fig. 10.** Reconstructed HR images (scaling factor 3) of *noisy Girl* by different methods. Top row: LR image, method [10] (PSNR=30.37dB, SSIM=0.7044) and method [47] (PSNR=29.77dB, SSIM=0.7258). Bottom row: method [43] (PSNR=31.40 dB, SSIM=0.7480), method [25] (PSNR=30.70dB, SSIM=0.7088) and proposed (PSNR=31.80dB, SSIM=0.7590).

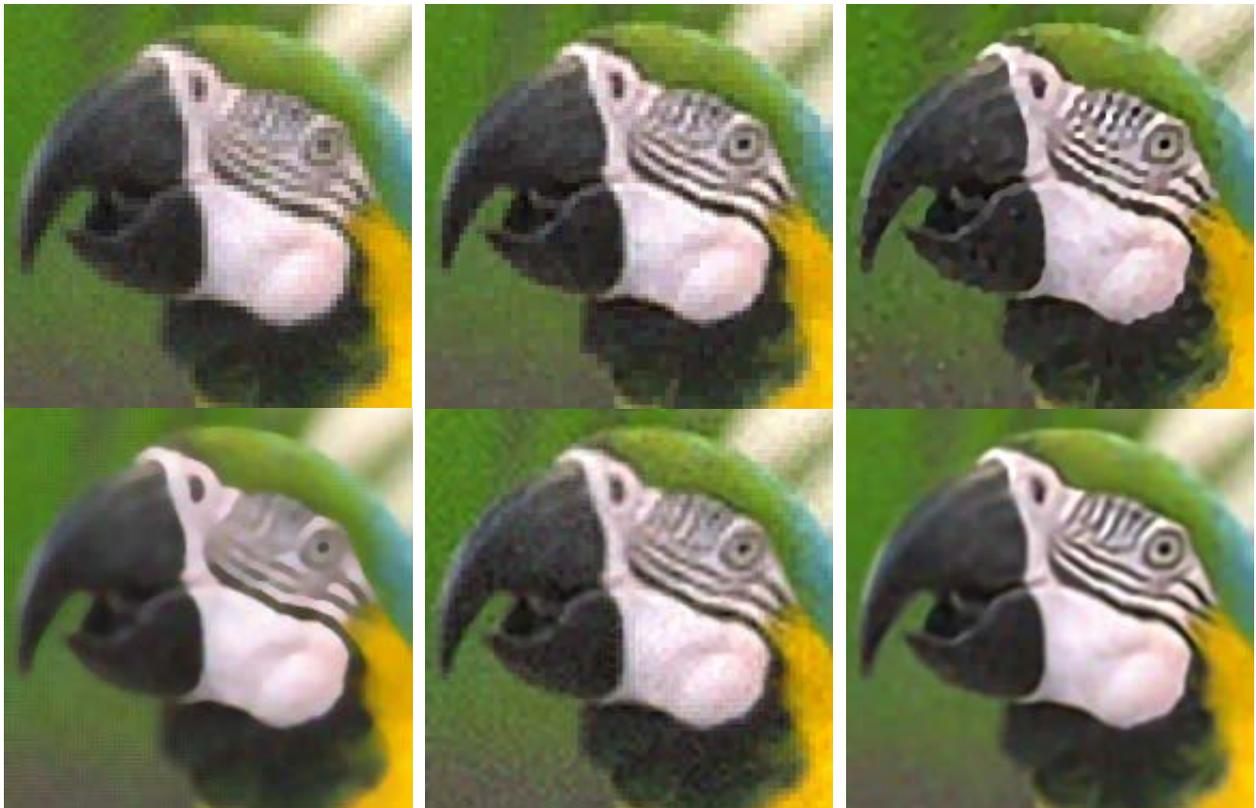

**Fig. 11.** Reconstructed HR images (scaling factor 3) of *noisy Parrot* by different methods. Top row: LR image, method [10] (PSNR=27.01dB, SSIM=0.7901) and method [47] (PSNR=26.77dB, SSIM=0.8084). Bottom row: method [43] (PSNR=27.42 dB, SSIM=0.8458), method [25] (PSNR=26.82dB, SSIM=0.7769) and proposed (PSNR=28.72dB, SSIM=0.8668).

In this section we present experimental results of single image super-resolution. Again we first test the robustness of the proposed method to the training dataset. Fig. 7 shows the reconstructed HR *Parrot* images by the proposed methods. We can see that the proposed method with the two different training datasets produces almost the same HR images. It can also be observed that the ASDS scheme can well reconstruct the image, while there are still some ringing artifacts around the reconstructed edges. Such artifacts can be reduced by coupling ASDS with the AR model based regularization, and the image quality can be further improved by incorporating the non-local similarity regularization.

Next we compare the proposed methods with state-of-the-art methods in [10, 43, 25, 47]. The visual comparisons are shown in Figs. 8~9. We see that the reconstructed HR images by method [10] have many jaggy and ringing artifacts. The TV-regularization based method [47] is effective in suppressing the ringing artifacts, but it generates piecewise constant block artifacts. The Softcuts method [43] produces very smooth edges and fine structures, making the reconstructed image look unnatural. By sparsely coding the LR image patches with the learned LR dictionary and recovering the HR image patches with the corresponding HR dictionary, the sparsity-based method in [25] is very competitive in terms of visual quality. However, it is difficult to learn a universal LR/HR dictionary pair that can represent various LR/HR structure pairs. It is observed that the reconstructed edges by [25] are relatively smooth and some fine image structures are not recovered. The proposed method generates the best visual quality. The reconstructed edges are much sharper than all the other four competing methods, and more image fine structures are recovered.

Often in practice the LR image will be noise corrupted, which makes the super-resolution more challenging. Therefore it is necessary to test the robustness of the super-resolution methods to noise. We added Gaussian white noise (with standard deviation 5) to the LR images, and the reconstructed HR images are shown in Figs. 10~11. We see that the method in [10] is sensitive to noise and there are serious noise-caused artifacts around the edges. The TV-regularization based method [47] also generates many noise-caused artifacts in the neighborhood of edges. The Softcuts method [43] results in over-smoothed HR images. Since the sparse representation based method [25] is followed by a back-projection process to remove the blurring effect, it is sensitive to noise and the performance degrades much in the noisy case. In contrast, the proposed method shows good robustness to noise. Not only the noise is effectively suppressed, but also the image fine edges are well reconstructed. This is mainly because the noise can be more effectively removed and the edges can be better preserved in the adaptive sparse domain. From Tables 5 and 6, we see that the average PSNR gains of ASDS-AR-NL-TD2 over the second best methods [10] (for the noiseless case) and [43] (for the noisy case) are 1.13 dB and 0.77 dB, respectively. The average SSIM gains over the methods [10] and [43] are 0.0348 and 0.021 for the noiseless and noisy cases, respectively.

## E. Experimental results on a 1000-image dataset

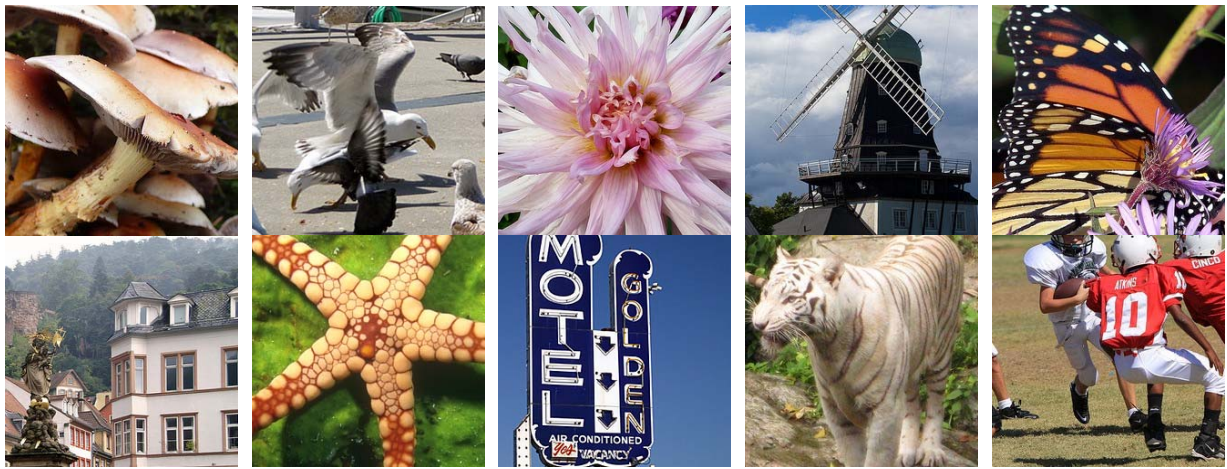

**Fig. 12.** Some example images in the established 1000-image dataset.

To more comprehensively test the robustness of the proposed IR method, we performed extensive deblurring and super-resolution experiments on a large dataset that contains 1000 natural images of various contents. To establish this dataset, we randomly downloaded 822 high-quality natural images from the *Flickr* website (http://www.flickr.com/), and selected 178 high-quality natural images from the *Berkeley Segmentation Database*[3]. A 256×256 sub-image that is rich in edge and texture structures was cropped from each of these 1000 images to test our method. Fig. 12 shows some example images in this dataset.

For image deblurring, we compared the proposed method with the methods in [46] and [58], which perform the 2[nd] and the 3[rd] best in our experiments in Section VI-D. The average PSNR and SSIM values of the deblurred images by the test methods are shown in Table 7. To better illustrate the advantages of the proposed method, we also drew the distributions of its PSNR gains over the two competing methods in Fig. 13. From Table 7 and Fig. 13, we can see that the proposed method constantly outperforms the competing methods for the uniform blur kernel, and the average PSNR gain over the BM3D [58] is up to 0.85 dB (when $\sigma_n=\sqrt{2}$). Although the performance gaps between different methods become much smaller for the non-truncated Gaussian blur kernel, it can still be observed that the proposed method mostly outperforms

---

[3] http://www.eecs.berkeley.edu/Research/Projects/CS/vision/grouping/segbench

BM3D [58] and [46], and the average PSNR gain over BM3D [58] is up to 0.19 dB (when $\sigma_n$=2). For image super-resolution, we compared the proposed method with the two methods in [25] and [47]. The average PSNR and SSIM values by the test methods are listed in Table 8, and the distributions of PSNR gain of our method over [25] and [47] are shown in Fig. 14. From Table 8 and Fig. 14, we can see that the proposed method performs constantly better than the competing methods.

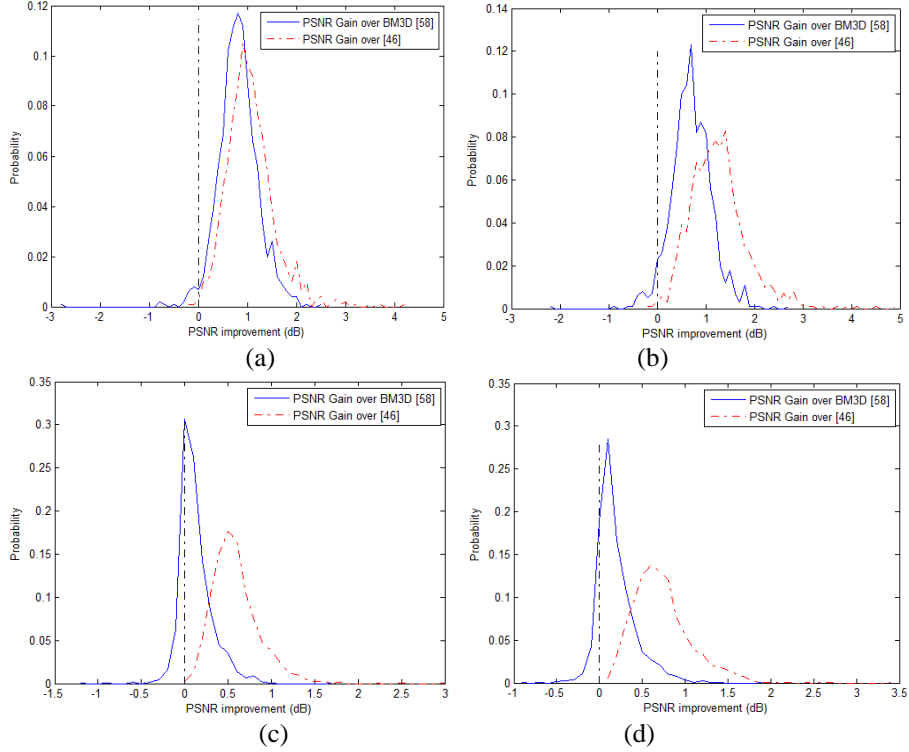

**Fig. 13.** The PSNR gain distributions of deblurring experiments. (a) Uniform blur kernel with $\sigma_n=\sqrt{2}$; (b) Uniform blur kernel with $\sigma_n$=2; (c) Gaussian blur kernel with $\sigma_n=\sqrt{2}$; (d) Gaussian blur kernel with $\sigma_n$=2.

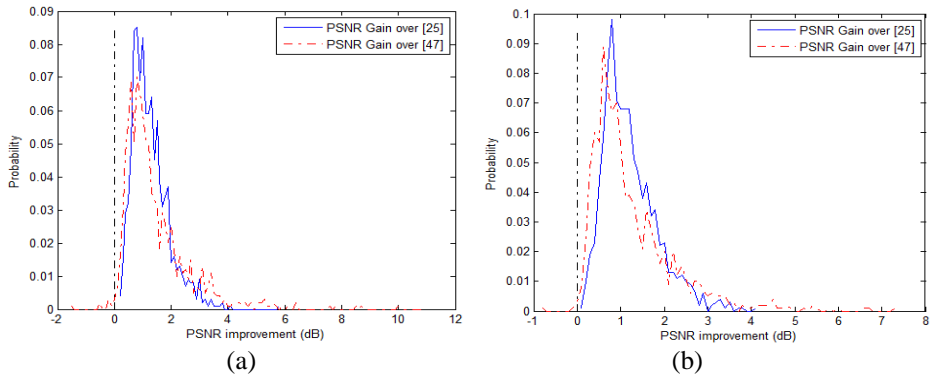

**Fig. 14.** The PSNR gain distributions of super-resolution experiments. (a) Noise level $\sigma_n$=0; (b) Noise level $\sigma_n$=5.

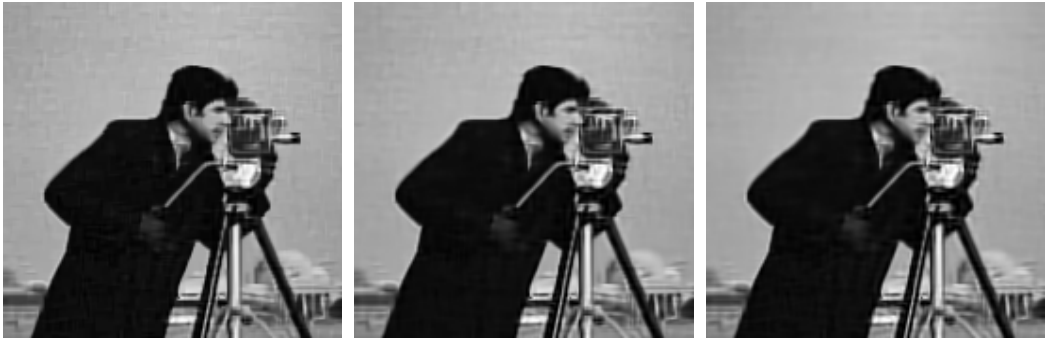

**Fig. 15.** Visual comparison of the deblurred images by the proposed method with different patch sizes. From left to right: patch size of 3×3, patch size of 5×5, and patch size of 7×7.

With this large dataset, we tested the robustness of the proposed method to the number of classes in learning the sub-dictionaries and AR models. Specifically, we trained the sub-dictionaries and AR models with different numbers of classes, i.e., 100, 200 and 400, and applied them to the established 1000-image dataset. Table 9 presents the average PSNR and SSIM values of the restored images. We can see that the three different numbers of classes lead to very similar image deblurring and super-resolution performance. This illustrates the robustness of the proposed method to the number of classes.

Another important issue of the proposed method is the size of image patch. Clearly, the patch size cannot be big; otherwise, they will not be micro-structures and hence cannot be represented by a small number of atoms. To evaluate the effects of the patch size on IR results, we trained the sub-dictionaries and AR models with different patch sizes, i.e., 3×3, 5×5 and 7×7. Then we applied these sub-dictionaries and AR models to the 10 test images and the constructed 1000-image database. The experimental results of deblurring and super-resolution are presented in Tables 10~12, from which we can see that these different patch sizes lead to similar PSNR and SSIM results. However, it can be found that the smaller patch sizes (i.e., 3×3 and 5×5) tend to generate some artifacts in smooth regions, as shown in Fig. 15. Therefore, we adopt 7×7 as the image patch size in our implementation.

## *F. Discussions on the computational cost*

In **Algorithm 1**, the matrices $U$ and $V$ are sparse matrices, and can be pre-calculated after the initialization of the AR models and the non-local weights. Hence, Step 2(a) can be executed fast. For image deblurring, the calculation of $U\hat{x}^{(k)}$ can be implemented by FFT, which is faster than direct matrix calculation. Steps 2(b) and 2(d) require $Nn^2$ multiplications, where $n$ is the number of pixels of each patch and $N$ is the number of patches. In our implementation, $N=N_I/4$, where $N_I$ is the number of pixels of the entire image. Since each patch can be sparsely coded individually, Steps 2(b) and 2(d) can be executed in parallel to speed up the algorithm. The update of sub-dictionaries and AR models requires $N$ operations of nearest neighbor search. We update them in every $P$ iterations ($P$=100 in our implementation) to speed up **Algorithm 1**. As an iterative shrinkage algorithm, the proposed **Algorithm 1** converges in 700~1000 iterations in most cases. For a 256×256 image, the proposed algorithm requires about 2~5 minutes for image deblurring and super-resolution on an Intel Core2 Duo 2.79G PC under the Matlab R2010a programming environment. In addition, several accelerating techniques, such as [51, 52], can be used to accelerate the convergence of the proposed algorithm. Hence, the computational cost of the proposed method can be further reduced.

## VII. Conclusion

We proposed a novel sparse representation based image deblurring and (single image) super-resolution method using adaptive sparse domain selection (ASDS) and adaptive regularization (AReg). Considering the fact that the optimal sparse domains of natural images can vary significantly across different images and different image patches in a single image, we selected adaptively the dictionaries that were pre-learned from a dataset of high quality example patches for each local patch. The ASDS improves significantly the effectiveness of sparse modeling and consequently the results of image restoration. To further improve the quality of reconstructed images, we introduced two AReg terms into the ASDS based image restoration framework. A set of autoregressive (AR) models were learned from the training dataset and were used to regularize the image local smoothness. The image non-local similarity was incorporated as another regularization term to exploit the image non-local redundancies. An iterated shrinkage algorithm was proposed to implement the proposed ASDS algorithm with AReg. The experimental results on natural images showed that the proposed ASDS-AReg approach outperforms many state-of-the-art methods in both PSNR and visual quality.

**Table 1.** PSNR (dB) and SSIM results of deblurred images (uniform blur kernel, noise level $\sigma_n=\sqrt{2}$).

| Images | [10] | [42] | [45] | [46] | [58] | ASDS-TD1 | ASDS-TD2 | ASDS-AR-TD1 | ASDS-AR-TD2 | ASDS-AR-NL-TD1 | ASDS-AR-NL-TD2 |
|---|---|---|---|---|---|---|---|---|---|---|---|
| Barbara | 25.83 0.7492 | 25.59 0.7373 | 26.11 0.7580 | 26.28 0.7671 | **27.90** 0.8171 | 26.60 0.7764 | 26.65 0.7709 | 26.93 0.7932 | 26.99 0.7893 | 27.63 0.8166 | 27.70 **0.8192** |
| Bike | 23.09 0.6959 | 24.24 0.7588 | 24.38 0.7564 | 24.15 0.7530 | 24.77 0.7740 | 25.29 0.8014 | 25.50 0.8082 | 25.21 0.7989 | 25.40 0.8052 | 25.32 0.8003 | **25.48** **0.8069** |
| Straw | 20.96 0.4856 | 21.31 0.5415 | 21.65 0.5594 | 21.32 0.5322 | **22.67** 0.6541 | 22.32 0.6594 | 22.38 **0.6651** | 22.39 0.6563 | 22.45 0.6615 | 22.51 0.6459 | 22.56 0.6540 |
| Boats | 28.80 0.8274 | 28.94 0.8331 | 29.44 0.8459 | 29.81 0.8496 | 29.90 0.8528 | 28.85 0.8076 | 28.94 0.8039 | 29.40 0.8286 | 29.48 0.8272 | 30.73 0.8665 | **30.76** **0.8670** |
| Parrots | 27.80 0.8652 | 28.80 0.8704 | 28.96 0.8722 | 29.04 0.8824 | 30.22 0.8906 | 30.71 0.8926 | 30.90 **0.8941** | 30.64 0.8920 | 30.79 0.8933 | 30.76 0.8921 | **30.92** 0.8939 |
| Baboon | 21.06 0.4811 | 21.16 0.5095 | 21.33 0.5192 | 21.21 0.5126 | 21.46 0.5315 | 21.43 **0.5881** | 21.45 0.5863 | 21.56 0.5878 | 21.55 0.5853 | 21.62 0.5754 | **21.62** 0.5765 |
| Hat | 29.75 0.8393 | 31.13 0.8624 | 30.88 0.8567 | 30.91 0.8591 | 30.85 0.8608 | 31.46 0.8702 | 31.67 **0.8736** | 31.41 0.8692 | 31.58 0.8721 | 31.43 0.8689 | **31.65** 0.8733 |
| Penta-gon | 24.69 0.6452 | 25.12 0.6835 | 25.57 0.7020 | 25.26 0.6830 | 26.00 0.7210 | 25.58 0.7285 | 25.62 0.7290 | 25.88 0.7385 | 25.89 0.7380 | 26.41 0.7511 | **26.46** **0.7539** |
| Camera-man | 25.73 0.8161 | 26.72 0.8330 | 27.38 0.8443 | 26.86 0.8361 | 27.24 0.8308 | 27.01 0.7956 | 27.14 0.7836 | 27.25 0.8255 | 27.37 0.8202 | 27.87 0.8578 | **28.00** **0.8605** |
| Peppers | 27.89 0.8123 | 28.44 0.8131 | 28.87 0.8298 | 28.75 0.8274 | 28.70 0.8151 | 28.24 0.7749 | 28.25 0.7682 | 28.64 0.7992 | 28.68 0.7941 | 29.46 0.8357 | **29.51** **0.8359** |
| Average | 25.56 0.7217 | 26.15 0.7443 | 26.46 0.7544 | 26.36 0.7500 | 26.97 0.7748 | 26.75 0.7695 | 26.85 0.7683 | 26.93 0.7789 | 27.02 0.7786 | 27.37 0.7910 | **27.47** **0.7943** |

**Table 2.** PSNR (dB) and SSIM results of deblurred images (uniform blur kernel, noise level $\sigma_n=2$).

| Images | [10] | [42] | [45] | [46] | [58] | ASDS-TD1 | ASDS-TD2 | ASDS-AR-TD1 | ASDS-AR-TD2 | ASDS-AR-NL-TD1 | ASDS-AR-NL-TD2 |
|---|---|---|---|---|---|---|---|---|---|---|---|
| Barbara | 24.86 0.6963 | 25.12 0.7031 | 25.34 0.7214 | 25.37 0.7248 | **27.16** 0.7881 | 26.33 0.7756 | 26.35 0.7695 | 26.45 0.7784 | 26.48 0.7757 | 26.89 0.7899 | 26.96 **0.7927** |
| Bike | 22.30 0.6391 | 24.07 0.7487 | 23.61 0.7142 | 23.33 0.7049 | 24.13 0.7446 | 24.46 0.7608 | 24.61 0.7670 | 24.43 0.7599 | 24.58 0.7656 | 24.59 0.7649 | **24.72** **0.7692** |
| Straw | 20.39 0.4112 | 21.07 0.5300 | 21.00 0.4885 | 20.81 0.4727 | **21.98** 0.5946 | 21.78 0.5991 | 21.78 **0.6027** | 21.79 0.5970 | 21.80 0.6008 | 21.81 0.5850 | 21.88 0.5934 |
| Boats | 27.47 0.7811 | 27.85 0.7880 | 28.66 0.8201 | 28.75 0.8181 | 29.19 0.8335 | 28.80 0.8145 | 28.83 0.8124 | 28.97 0.8195 | 29.00 0.8187 | **29.83** **0.8441** | 29.83 0.8435 |
| Parrots | 26.84 0.8432 | 28.58 0.8595 | 28.06 0.8573 | 27.98 0.8665 | 29.45 0.8806 | 29.77 0.8787 | 29.98 0.8802 | 29.73 0.8784 | 29.94 0.8798 | 29.94 0.8800 | **30.06** **0.8807** |
| Baboon | 20.58 0.4048 | 20.98 0.4965 | 20.87 0.4528 | 20.80 0.4498 | 21.13 0.4932 | 21.10 **0.5441** | 21.10 0.5429 | 21.17 0.5428 | 21.16 0.5410 | 21.24 0.5285 | **21.24** 0.5326 |
| Hat | 28.92 0.8153 | 30.79 0.8524 | 30.28 0.8433 | 30.15 0.8420 | 30.36 0.8507 | 30.71 0.8522 | 30.89 0.8556 | 30.69 0.8516 | 30.86 0.8550 | 30.80 0.8545 | **30.99** **0.8574** |
| Penta-gon | 23.88 0.5776 | 24.59 0.6587 | 24.86 0.6516 | 24.54 0.6297 | 25.46 0.6885 | 25.34 0.7051 | 25.31 0.7042 | 25.42 0.7069 | 25.39 0.7066 | 25.74 0.7118 | **25.75** **0.7146** |
| Camera-man | 24.80 0.7837 | 26.04 0.7772 | 26.53 0.8273 | 25.96 0.8131 | 26.53 0.8136 | 26.67 0.8211 | 26.81 0.8156 | 26.69 0.8243 | 26.86 0.8238 | 27.11 0.8365 | **27.25** **0.8408** |
| Peppers | 27.04 0.7889 | 27.46 0.7660 | 28.33 0.8144 | 28.05 0.8106 | 28.15 0.7999 | 28.30 0.7995 | 28.24 0.7904 | 28.37 0.8038 | 28.37 0.7988 | 28.82 0.8204 | **28.87** **0.8209** |
| Average | 24.71 0.6741 | 25.66 0.7180 | 25.75 0.7191 | 25.57 0.7132 | 26.35 0.7487 | 26.33 0.7551 | 26.39 0.7540 | 26.37 0.7562 | 26.44 0.7566 | 26.68 0.7615 | **26.75** **0.7646** |

**Table 3.** PSNR (dB) and SSIM results of deblurred images (Gaussian blur kernel, noise level $\sigma_n=\sqrt{2}$).

| Images | [10] | [42] | [45] | [46] | [58] | ASDS-TD1 | ASDS-TD2 | ASDS-AR-TD1 | ASDS-AR-TD2 | ASDS-AR-NL-TD1 | ASDS-AR-NL-TD2 |
|---|---|---|---|---|---|---|---|---|---|---|---|
| *Barbara* | 23.65 0.6411 | 23.22 0.5971 | 23.19 0.5892 | 23.71 0.6460 | 23.77 0.6489 | 23.81 0.6560 | 23.81 0.6556 | 23.81 0.6566 | 23.81 0.6563 | 23.86 0.6609 | **23.86 0.6611** |
| *Bike* | 21.78 0.6085 | 21.90 0.6137 | 21.20 0.5515 | 22.20 0.6407 | 22.71 0.6774 | 22.59 0.6657 | 22.63 0.6693 | 22.59 0.6663 | 22.62 0.6688 | 22.80 0.6813 | **22.82 0.6830** |
| *Straw* | 20.28 0.4005 | 19.76 0.3502 | 19.33 0.2749 | 20.33 0.4087 | **21.02 0.5003** | 20.76 0.4710 | 20.81 0.4754 | 20.79 0.4729 | 20.82 0.4773 | 20.91 0.4866 | 20.93 0.4894 |
| *Boats* | 26.19 0.7308 | 25.53 0.7056 | 24.77 0.6688 | 26.64 0.7464 | 26.99 0.7486 | 27.12 0.7617 | 27.14 0.7633 | 27.11 0.7616 | 27.13 0.7625 | 27.27 0.7651 | **27.31 0.7677** |
| *Parrots* | 26.40 0.8321 | 25.96 0.8080 | 25.21 0.7949 | 26.84 0.8444 | **27.72** 0.8580 | 27.42 0.8539 | 27.50 0.8538 | 27.45 0.8540 | 27.52 0.8540 | 27.67 **0.8600** | 27.70 0.8598 |
| *Baboon* | 20.22 0.3622 | 20.01 0.3396 | 19.85 0.3011 | 20.24 0.3673 | 20.34 0.3923 | 20.36 0.3908 | 20.35 0.3889 | 20.36 0.3916 | 20.35 0.3893 | **20.39 0.3976** | 20.38 0.3959 |
| *Hat* | 28.11 0.7916 | 28.90 0.8100 | 28.29 0.7924 | 28.85 0.8122 | 28.87 0.8119 | 28.80 0.8074 | 28.92 0.8104 | 28.80 0.8074 | 28.89 0.8099 | 28.96 0.8110 | **29.01 0.8134** |
| *Penta-gon* | 23.33 0.5472 | 22.48 0.4881 | 22.09 0.4387 | 23.39 0.5540 | 23.82 0.5994 | 23.89 0.5974 | 23.88 0.5958 | 23.89 0.5978 | 23.89 0.5971 | 24.00 0.6086 | **24.01 0.6089** |
| *Camera-man* | 23.08 0.7332 | 23.26 0.7483 | 22.59 0.7187 | 23.51 0.7521 | 23.77 0.7249 | 23.85 0.7603 | 23.90 0.7637 | 23.83 0.7599 | 23.89 0.7630 | 24.03 0.7619 | **24.05 0.7649** |
| *Peppers* | 25.96 0.7666 | 25.58 0.7411 | 24.94 0.7236 | 26.61 0.7843 | 26.65 0.7626 | 26.99 0.7883 | 27.01 0.7900 | 26.98 0.7880 | 26.99 0.7898 | 27.12 0.7880 | **27.14 0.7902** |
| *Average* | 23.90 0.6414 | 23.66 0.6202 | 23.15 0.5854 | 24.23 0.6556 | 24.57 0.6724 | 24.56 0.6752 | 24.59 0.6766 | 24.56 0.6756 | 24.59 0.6768 | 24.70 0.6821 | **24.72 0.6834** |

**Table 4.** PSNR (dB) and SSIM results of deblurred images (Gaussian blur kernel, noise level $\sigma_n=2$).

| Images | [10] | [42] | [45] | [46] | [58] | ASDS-TD1 | ASDS-TD2 | ASDS-AR-TD1 | ASDS-AR-TD2 | ASDS-AR-NL-TD1 | ASDS-AR-NL-TD2 |
|---|---|---|---|---|---|---|---|---|---|---|---|
| *Barbara* | 23.57 0.6309 | 23.19 0.5933 | 23.07 0.5776 | 23.62 0.6351 | 23.70 0.6399 | 23.72 0.6464 | 23.72 0.6464 | 23.73 0.6468 | 23.73 0.6471 | 23.78 0.6520 | **23.78 0.6521** |
| *Bike* | 21.58 0.5903 | 21.88 0.6125 | 20.97 0.5324 | 21.93 0.6178 | 22.53 0.6643 | 22.41 0.6506 | 22.45 0.6527 | 22.41 0.6513 | 22.45 0.6536 | 22.66 0.6685 | **22.69 0.6704** |
| *Straw* | 20.10 0.3750 | 19.75 0.3499 | 19.24 0.2590 | 20.10 0.3781 | **20.81 0.4762** | 20.57 0.4471 | 20.60 0.4500 | 20.58 0.4484 | 20.62 0.4529 | 20.72 0.4664 | 20.75 0.4698 |
| *Boats* | 25.87 0.7157 | 25.48 0.7032 | 24.63 0.6602 | 26.24 0.7292 | 26.71 0.7359 | 26.78 0.7464 | 26.82 0.7488 | 26.81 0.7478 | 26.81 0.7487 | 26.98 0.7503 | **26.96 0.7521** |
| *Parrots* | 26.10 0.8234 | 25.92 0.8053 | 25.05 0.7907 | 26.38 0.8337 | 27.40 0.8523 | 27.08 0.8443 | 27.14 0.8447 | 27.13 0.8452 | 27.24 0.8460 | 27.47 **0.8536** | **27.50** 0.8535 |
| *Baboon* | 20.16 0.3497 | 20.00 0.3389 | 19.79 0.2905 | 20.17 0.3533 | 20.28 0.3826 | 20.28 0.3775 | 20.28 0.3758 | 20.29 0.3775 | 20.28 0.3762 | **20.32 0.3858** | 20.31 0.3839 |
| *Hat* | 27.94 0.7857 | 28.86 **0.8084** | 28.27 0.7913 | 28.59 0.8043 | 28.67 0.8049 | 28.59 0.8009 | 28.69 0.8036 | 28.59 0.8009 | 28.69 0.8036 | 28.80 0.8056 | **28.87** 0.8080 |
| *Penta-gon* | 23.13 0.5267 | 22.46 0.4876 | 21.89 0.4200 | 23.13 0.5299 | 23.65 0.5843 | 23.69 0.5784 | 23.69 0.5770 | 23.69 0.5793 | 23.70 0.5783 | 23.80 **0.5922** | **23.81** 0.5917 |
| *Camera-man* | 22.93 0.7256 | 23.23 0.7465 | 22.36 0.7130 | 23.25 0.7412 | 23.60 0.7198 | 23.72 0.7533 | 23.76 0.7568 | 23.71 0.7528 | 23.76 0.7564 | 23.95 0.7557 | **23.95 0.7583** |
| *Peppers* | 25.72 0.7570 | 25.50 0.7373 | 24.38 0.7034 | 26.24 0.7723 | 26.44 0.7555 | 26.70 0.7770 | 26.76 0.7800 | 26.71 0.7773 | 26.76 0.7804 | 26.91 0.7774 | **26.93 0.7799** |
| *Average* | 23.71 0.6280 | 23.63 0.6183 | 22.96 0.5738 | 23.97 0.6395 | 24.38 0.6616 | 24.36 0.6622 | 24.39 0.6636 | 24.37 0.6627 | 24.40 0.6643 | 24.54 0.6707 | **24.56 0.6720** |

**Table 5.** The PSNR (dB) and SSIM results (luminance components) of reconstructed HR images (noise level $\sigma_n$=0).

| Images | [10] | [43] | [25] | [47] | ASDS-TD1 | ASDS-TD2 | ASDS-AR-TD1 | ASDS-AR-TD2 | ASDS-AR-NL-TD1 | ASDS-AR-NL-TD2 |
|---|---|---|---|---|---|---|---|---|---|---|
| *Girl* | 32.93 0.8102 | 31.94 0.7704 | 32.51 0.7912 | 31.21 0.7878 | 33.40 0.8213 | 33.41 0.8215 | 33.42 0.8218 | 33.41 0.8216 | **33.54** **0.8242** | 33.53 0.8242 |
| *Parrot* | 28.78 0.8845 | 27.71 0.8682 | 27.98 0.8665 | 27.59 0.8856 | 29.47 0.9031 | 29.51 0.9034 | 29.61 0.9036 | 29.63 0.9038 | 29.97 0.9090 | **30.00** **0.9093** |
| *Butterfly* | 25.16 0.8336 | 25.19 0.8623 | 23.73 0.7942 | 26.60 0.9036 | 26.24 0.8775 | 26.27 0.8779 | 26.24 0.8758 | 26.23 0.8753 | 27.09 0.8975 | **27.34** **0.9047** |
| *Leaves* | 24.59 0.8310 | 24.34 0.8372 | 24.35 0.8170 | 24.58 0.8878 | 25.94 0.8847 | 25.97 0.8856 | 25.93 0.8835 | 25.95 0.8842 | 26.78 0.9050 | **26.80** **0.9058** |
| *Parthenon* | 26.32 0.7135 | 25.87 0.6791 | 24.08 0.6305 | 25.89 0.7163 | 26.63 0.7279 | 26.61 0.7278 | 26.63 0.7279 | 26.62 0.7277 | 26.82 0.7348 | **26.83** **0.7349** |
| *Flower* | 28.16 0.8120 | 27.50 0.7800 | 27.76 0.7929 | 27.38 0.8111 | 28.80 0.8351 | 28.82 0.8354 | 28.82 0.8352 | 28.84 0.8358 | 29.19 0.8480 | **29.19** **0.8480** |
| *Hat* | 29.92 0.8438 | 29.68 0.8389 | 29.65 0.8362 | 29.19 0.8569 | 30.70 0.8653 | 30.69 0.8648 | 30.65 0.8643 | 30.64 0.8641 | 30.92 **0.8707** | **30.93** 0.8706 |
| *Raccoon* | 28.80 0.7549 | 27.96 0.6904 | 28.49 0.7273 | 27.53 0.7076 | 29.06 0.7648 | 29.10 0.7658 | 29.11 0.7657 | 29.13 0.7664 | 29.23 0.7675 | **29.24** **0.7677** |
| *Bike* | 23.48 0.7438 | 23.31 0.7219 | 23.20 0.7188 | 23.61 0.7567 | 24.10 0.7760 | 24.11 0.7772 | 24.08 0.7752 | 24.07 0.7752 | 24.48 0.7948 | **24.62** **0.7962** |
| *Plants* | 31.87 0.8792 | 31.45 0.8617 | 31.48 0.8698 | 31.28 0.8784 | 32.85 0.8985 | 32.91 0.8996 | 32.85 0.8987 | 32.88 0.8995 | 33.47 0.9094 | **33.47** **0.9095** |
| *Average* | 28.03 0.8115 | 27.49 0.7910 | 27.69 0.7954 | 27.49 0.8190 | 28.72 0.8354 | 28.74 0.8359 | 28.73 0.8352 | 28.74 0.8354 | 29.15 0.8461 | **29.16** **0.8463** |

**Table 6.** The PSNR (dB) and SSIM results (luminance components) of reconstructed HR images (noise level $\sigma_n$=5).

| Images | [10] | [43] | [25] | [47] | ASDS-TD1 | ASDS-TD2 | ASDS-AR-TD1 | ASDS-AR-TD2 | ASDS-AR-NL-TD1 | ASDS-AR-NL-TD2 |
|---|---|---|---|---|---|---|---|---|---|---|
| *Noisy Girl* | 30.37 0.7044 | 31.40 0.7480 | 30.70 0.7088 | 29.77 0.7258 | 31.72 0.7583 | 31.76 **0.7596** | 31.72 0.7584 | 31.75 0.7594 | 31.79 0.7593 | **31.80** 0.7590 |
| *Noisy Parrot* | 27.01 0.7911 | 27.42 0.8458 | 26.82 0.7769 | 26.77 0.8084 | 28.81 0.8673 | **28.91** **0.8689** | 28.74 0.8634 | 28.83 0.8676 | 28.66 0.8632 | 28.72 0.8668 |
| *Noisy Butterfly* | 23.67 0.7777 | 24.95 0.8427 | 23.50 0.7576 | 25.47 0.8502 | 25.54 0.8362 | 25.76 0.8435 | 25.50 0.8350 | 25.61 0.8388 | 25.99 0.8591 | **26.08** **0.8612** |
| *Noisy Leaves* | 23.62 0.7751 | 23.17 0.7939 | 23.35 0.7467 | 23.78 0.8457 | 25.14 0.8457 | 25.21 0.8491 | 25.11 0.8444 | 25.13 0.8455 | 25.49 0.8633 | **25.50** **0.8645** |
| *Noisy Parthenon* | 25.31 0.6163 | 25.65 0.6587 | 23.89 0.5847 | 25.24 0.6651 | 26.06 0.6826 | 26.09 **0.6845** | 26.06 0.6816 | 26.08 0.6826 | 26.09 0.6807 | **26.10** 0.6821 |
| *Noisy Flower* | 26.61 0.6991 | 27.16 0.7591 | 26.51 0.7020 | 26.45 0.7509 | 27.58 0.7683 | 27.55 0.7699 | 27.64 0.7710 | 27.65 0.7733 | 27.67 0.7738 | **27.69** **0.7767** |
| *Noisy Hat* | 28.14 0.6944 | 29.27 0.8049 | 28.32 0.7282 | 28.11 0.7768 | 29.56 0.8086 | **29.70** 0.8151 | 29.50 0.8075 | 29.58 0.8129 | 29.57 0.8127 | 29.63 **0.8175** |
| *Noisy Raccoon* | 27.05 0.6434 | 27.60 0.6707 | 27.20 0.6418 | 26.73 0.6640 | 27.98 0.6886 | **28.01** **0.6882** | 27.99 0.6880 | 28.01 0.6876 | 28.01 0.6840 | 28.01 0.6810 |
| *Noisy Bike* | 22.74 0.6672 | 23.06 0.6984 | 22.42 0.6459 | 23.07 0.7118 | 23.49 0.7201 | **23.57** **0.7239** | 23.43 0.7182 | 23.49 0.7205 | 23.52 0.7205 | 23.57 0.7220 |
| *Noisy Plants* | 29.93 0.7760 | 30.80 0.8343 | 29.51 0.7691 | 29.67 0.8028 | 31.01 0.8324 | 31.03 0.8342 | 30.95 0.8308 | 30.99 0.8327 | 31.09 0.8350 | **31.10** **0.8363** |
| *Average* | 26.49 0.7048 | 27.05 0.7657 | 26.34 0.7090 | 26.52 0.7604 | 27.69 0.7808 | 27.76 0.7837 | 27.66 0.7798 | 27.71 0.7821 | 27.79 0.7851 | **27.82** **0.7867** |

**Table 7.** Average PSNR and SSIM values of the deblurred images on the 1000-image dataset.

| Method | Uniform blur kernel $\sigma_n=\sqrt{2}$ | Uniform blur kernel $\sigma_n=2$ | Gaussian blur kernel $\sigma_n=\sqrt{2}$ | Gaussian blur kernel $\sigma_n=2$ |
|---|---|---|---|---|
| ASDS-AR-NL-TD2 | **29.36 (0.8397)** | **28.66 (0.8163)** | **26.22 (0.7335)** | **26.10 (0.7261)** |
| [58] | 28.51 (0.8139) | 27.96 (0.7966) | 26.09 (0.7297) | 25.91 (0.7209) |
| [46] | 28.26 (0.8081) | 27.41 (0.7763) | 25.63 (0.7072) | 25.37 (0.6934) |

**Table 8.** Average PSNR and SSIM results of the reconstructed HR images on the 1000-image dataset.

| Method | Noise level $\sigma_n=0$ | Noise level $\sigma_n=5$ |
|---|---|---|
| ASDS-AR-NL-TD2 | **27.53 (0.7975)** | **26.56 (0.7444)** |
| [25] | 26.26 (0.7444) | 25.34 (0.6711) |
| [47] | 26.09 (0.7705) | 25.31 (0.7156) |

**Table 9.** Average PSNR and SSIM results by the proposed ASDS-AR-NL-TD2 method with different numbers of classes on the 1000-image dataset.

| Number of classes | Deblurring with uniform blur kernel and $\sigma_n=\sqrt{2}$ | Super-resolution with noise level $\sigma_n=0$ |
|---|---|---|
| 100 | 29.29 (0.8379) | 27.51 (0.7971) |
| 200 | **29.36 (0.8397)** | 27.52 (0.7974) |
| 400 | 29.31 (0.8380) | **27.53 (0.7975)** |

**Table 10.** The PSNR and SSIM results of deblurred images by the proposed ASDS-AR-NL-TD2 with different patch sizes (uniform blurring kernel, $\sigma_n=\sqrt{2}$).

| Patch Size | Barbara | Bike | Straw | Boats | Parrots | Baboon | Hat | Penta-gon | Camer-aman | Peppers | Average |
|---|---|---|---|---|---|---|---|---|---|---|---|
| 3×3 | 27.33 0.7936 | 25.68 0.8173 | 22.32 0.6320 | 30.64 0.8651 | 31.07 0.9024 | 21.61 0.5713 | 32.12 0.8816 | 26.44 0.7509 | 28.09 0.8455 | 29.55 0.8270 | 27.49 0.7887 |
| 5×5 | 27.59 0.8116 | 25.54 0.8089 | 22.44 0.6428 | 30.81 0.8689 | 31.04 0.8968 | 21.61 0.5751 | 31.84 0.8745 | 26.48 0.7549 | 28.11 0.8599 | 29.63 0.8339 | 27.51 0.7927 |
| 7×7 | 27.70 0.8192 | 25.48 0.8069 | 22.56 0.6540 | 30.76 0.8670 | 30.92 0.8939 | 21.62 0.5765 | 31.65 0.8733 | 26.46 0.7553 | 28.00 0.8605 | 29.51 0.8359 | 27.47 0.7943 |

**Table 11.** The PSNR and SSIM results of reconstructed HR images by the proposed ASDS-AR-NL-TD2 with different patch sizes (noise level $\sigma_n=0$).

| Patch Size | Girl | Parrot | Butterfly | Leaves | Parthenon | Flower | Hat | Raccoon | Bike | Plants | Average |
|---|---|---|---|---|---|---|---|---|---|---|---|
| 3×3 | 33.55 0.8251 | 29.96 0.9104 | 27.28 0.9055 | 27.00 0.9139 | 26.84 0.7366 | 29.27 0.8527 | 30.95 0.8739 | 29.18 0.7660 | 24.46 0.7961 | 33.54 0.9131 | 29.20 0.8493 |
| 5×5 | 33.56 0.8240 | 30.09 0.9121 | 27.39 0.9058 | 27.00 0.9118 | 26.90 0.7377 | 29.25 0.8500 | 31.10 0.8742 | 29.22 0.7664 | 24.53 0.7965 | 33.59 0.9116 | 29.26 0.8490 |
| 7×7 | 33.55 0.8204 | 30.14 0.9092 | 27.34 0.9047 | 26.93 0.9099 | 26.89 0.7357 | 29.19 0.8463 | 31.04 0.8716 | 29.24 0.7655 | 24.62 0.7962 | 33.37 0.9061 | 29.22 0.8464 |

**Table 12.** Average PSNR and SSIM results by the proposed ASDS-AR-NL-TD2 method with different patch sizes on the 1000-image dataset.

| Patch size | Deblurring with uniform blur kernel and $\sigma_n=\sqrt{2}$ | Super-resolution with noise level $\sigma_n=0$ |
|---|---|---|
| 3×3 | **29.60 (0.8466)** | 27.51 (0.7979) |
| 5×5 | 29.56 (0.8450) | **27.54 (0.7984)** |
| 7×7 | 29.36 (0.8397) | 27.53 (0.7976) |